\documentclass[preprint,12pt,authoryear]{elsarticle}

\usepackage{amssymb}

\usepackage{float}
\usepackage{placeins}
\usepackage{multirow}
\usepackage{multicol}
\usepackage{amsmath}
\usepackage{amssymb}
\usepackage{yhmath}
\usepackage{mathtools}
\usepackage{amsthm}
\usepackage{graphicx}
\usepackage{adjustbox}
\usepackage{listings}
\usepackage{hyperref}
\usepackage{array}
\usepackage[normalem]{ulem}
\usepackage{tabularx}
\usepackage{soul}

\usepackage{xcolor}

\newtheoremstyle{colon}%
{}
{}
{\itshape}
{}
{\bfseries}
{:}
{ }
{}
\theoremstyle{colon}
\newtheorem{problem}{Problem}
\newtheorem{proposition}{Proposition}

\newcommand{\bigcomment}[1]{\iffalse #1 \fi}

\newenvironment{rev}[1][\unskip]
{%
}
{
}

\let\oldcite\cite
\renewcommand{\cite}[1]{(\oldcite{#1})}

\journal{Artificial Intelligence}

\begin{document}

\begin{frontmatter}



\title{Integrating Symbolic Reasoning into Neural Generative Models for Design Generation}


\author{Maxwell J. Jacobson}
\author{Yexiang Xue$^*$}

\affiliation{organization={Purdue University, Department of Computer Science},
            addressline={305 N. University Street}, 
            city={West Lafayette},
            postcode={47907}, 
            state={IN},
            country={USA.},
            email={~~\{jacobs57, yexiang\}@purdue.edu}
            }

\begin{abstract}

Design generation requires tight integration of neural and symbolic reasoning, as good design must meet explicit user needs and honor implicit rules for aesthetics, utility, and convenience. 
Current automated design tools driven by neural networks produce appealing designs, but cannot satisfy user specifications and utility requirements. 
Symbolic reasoning tools, such as constraint programming, cannot perceive low-level visual information in images or capture subtle aspects such as aesthetics. 
We introduce the Spatial Reasoning Integrated Generator (SPRING) for design generation.
SPRING embeds a neural and symbolic integrated spatial reasoning module inside the deep generative network.
The spatial reasoning module samples the set of locations of objects to be generated from a backtrack-free distribution. 
This distribution modifies the implicit preference distribution, which is learned by a recursive neural network to capture utility and aesthetics. 
The sampling from the backtrack-free distribution is accomplished by a symbolic reasoning approach, SampleSearch, which zeros out the probability of sampling spatial locations violating explicit user specifications.
%
%
Embedding symbolic reasoning into neural generation guarantees that the output of SPRING satisfies user requirements. 
Furthermore, SPRING offers interpretability, allowing users to visualize and diagnose the generation process through the bounding boxes.
SPRING is also adept at managing novel user specifications not encountered during its training, thanks to its proficiency in zero-shot constraint transfer.
Quantitative evaluations and a human study reveal that SPRING outperforms baseline generative models, excelling in delivering high design quality and better meeting user specifications.

%

\end{abstract}

\begin{graphicalabstract}

\includegraphics[width=\textheight, height=0.75\textwidth, angle=270]{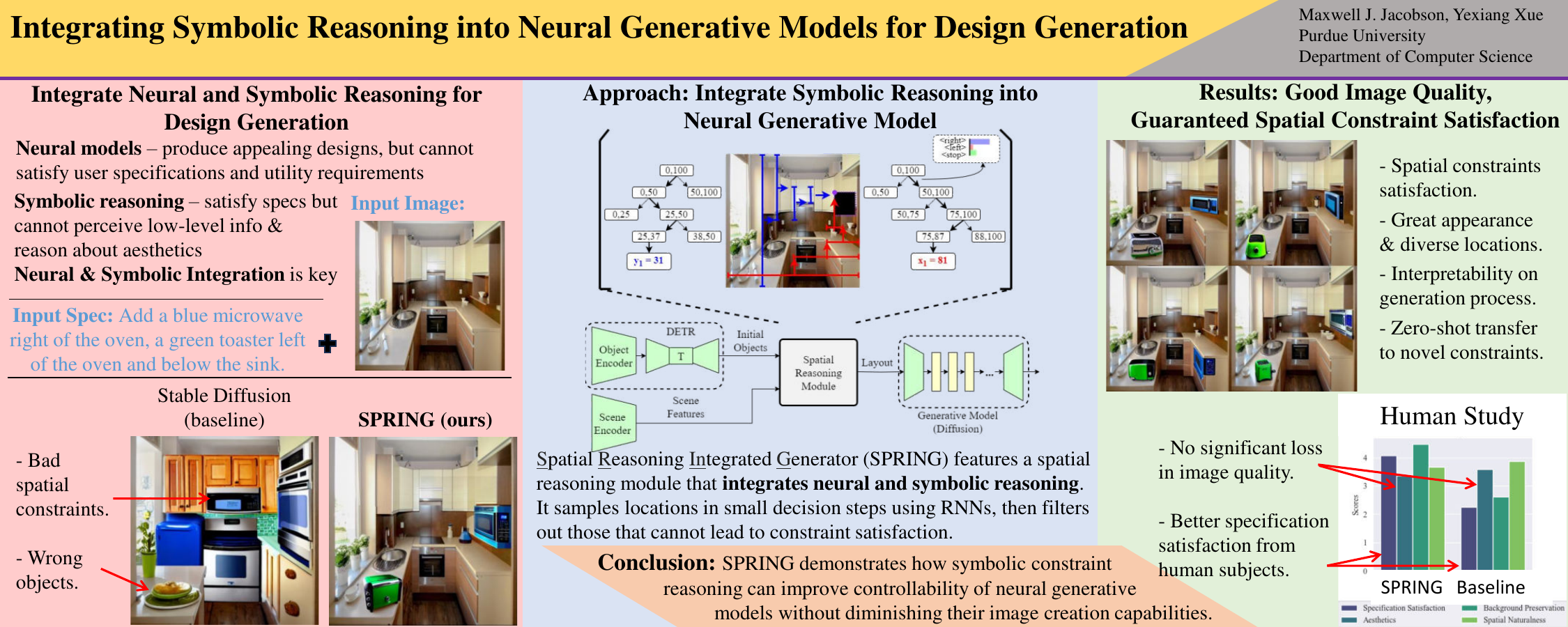}

\end{graphicalabstract}

\begin{highlights}
\item SPRING integrates symbolic and neural reasoning for constrained interior design
\item Constraint reasoning guarantees specification satisfaction; neural net paints realistic objects
\item Delivers high-quality design images aligned with user specifications and guarantees
\item Allows zero-shot transfer to new spatial constraints without re-training
\end{highlights}


\end{frontmatter}


\FloatBarrier
\section{Introduction}

\begin{figure}[t]
    \centering
    \includegraphics[width=\linewidth]{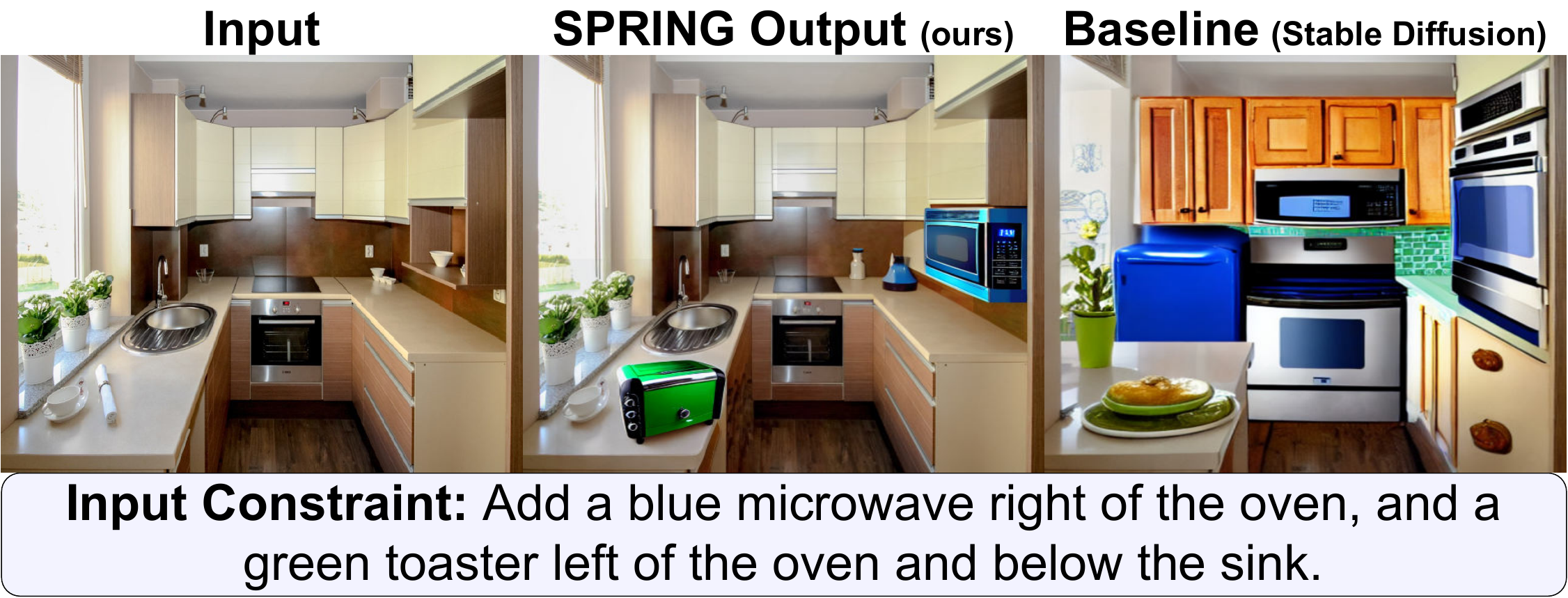}
    \vspace{-8pt}
    \caption{
    An interior design generated by our proposed SPRING model (middle) with a given background already containing an oven and a sink among other objects (left). The user specifications are at the bottom (provided to SPRING in the form of propositional logic; natural language text is used here to aid readability). SPRING creates a design satisfying the specifications. Text-to-image approaches like Stable Diffusion (right) often fail to meet these constraints, mixing up the number, color, and placement of objects.
    }
    \label{fig:functionality}
\end{figure}

\begin{figure*}[p]
    \centering
    \includegraphics[width=\textwidth]{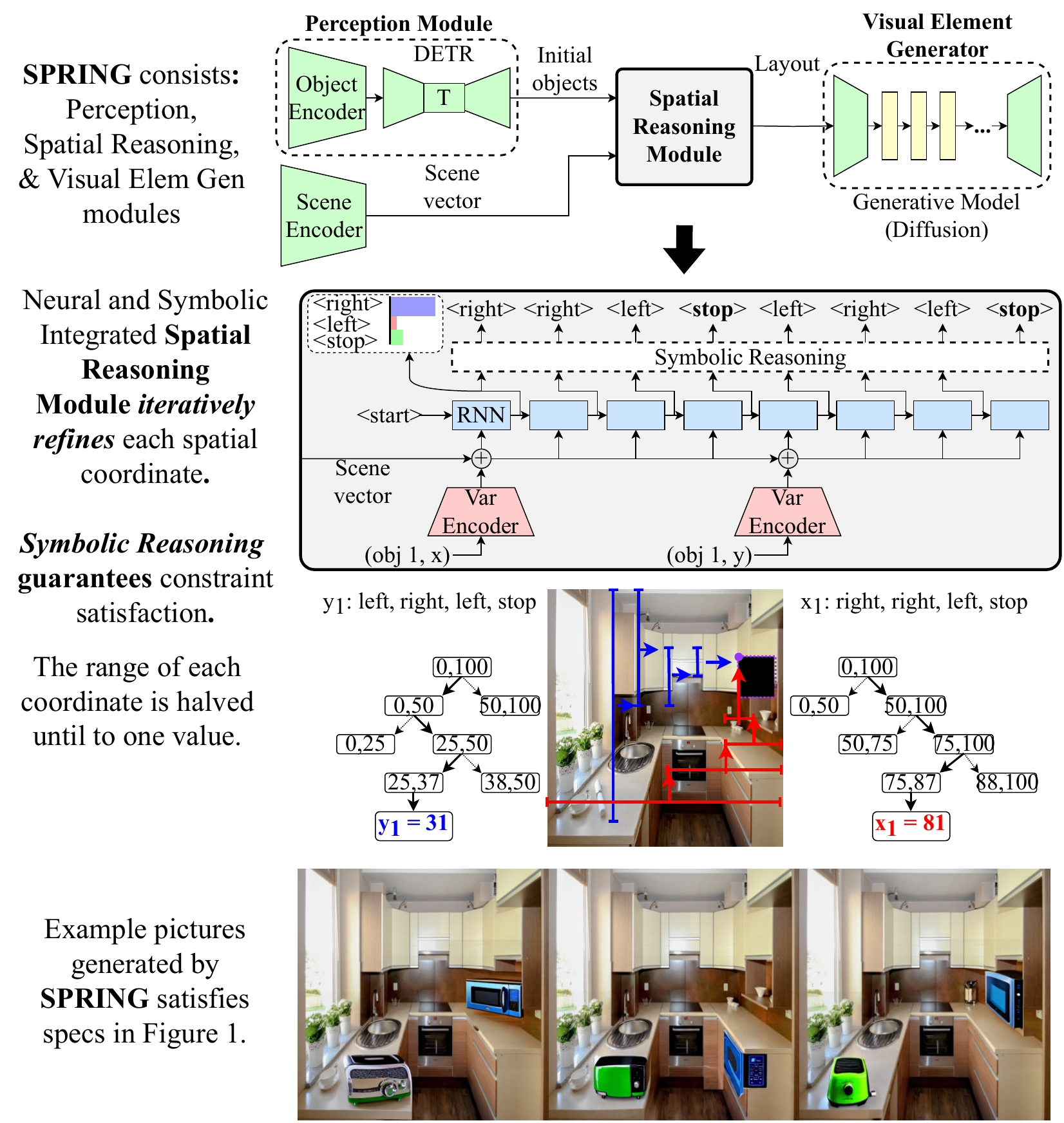}
    \caption{Rollout of the SPRING system. \textbf{(First row)} SPRING consists of the perception module, the spatial reasoning module, and the visual element generation module. \textbf{(Second \& third rows)} The neural and symbolic integrated spatial reasoning module decides the bounding boxes of each object to be generated. It 
    \textit{iteratively halves} each coordinate of every bounding box until it is sufficiently small. Symbolic reasoning in the form of SampleSearch is applied to the output of the neural net to ensure satisfaction of user constraints. 
 These bounding boxes are filled by the visual element generator in the last step.
 \textbf{(Fourth row)} Example pictures generated by SPRING demonstrate good quality designs satisfying user specifications in Figure \ref{fig:functionality}.
    }
    \label{fig:rollout}
    \vspace{-.3cm}
\end{figure*}

Neural and symbolic approaches are two fundamental drivers behind the progress of artificial intelligence (AI).  
Neural approaches have spearheaded major developments in learning from diverse and unstructured data, discovering hidden or fuzzy patterns, and producing effective predictive and generative models. Nevertheless, it is challenging for neural models to provide formal guarantees. Furthermore, predictions made by neural models may violate constraints, especially from rare and unseen inputs. 
Symbolic approaches have produced efficient and reliable algorithms that can provide formal guarantees, interpretability, and robustness. 
However, symbolic approaches often necessitate fully formulated problems prior to attempting solutions, leading to rigid models that are difficult to adapt to changing data distributions.
Neither neural nor symbolic methods offer a complete solution on their own, as each has its own set of strengths and weaknesses. Consequently, many meaningful real-world tasks remain outside the reach of current artificial intelligences. 
A better solution would be the \textit{integration} of neural and symbolic approaches.
However, it is not trivial to bridge the two approaches due to multiple aspects of incompatibility. 
Recently, exciting work has emerged to unite neural and symbolic reasoning 
\cite{kautz2022third,Mitchener2022,wu2023spring,mitchell2021conversational,NEURIPS2018_5e388103,NEURIPS2021_d3e2e8f6,mao2018the,745}. However, the complete success of neural-symbolic integration is far from achieved.

Design generation is a field that requires \textit{both} neural and symbolic reasoning. 
Good design must meet industry standards and user specifications while capturing subtle aspects such as aesthetics and convenience. 
Symbolic approaches -- for example, a constraint program -- can be defined to create a plan that meets all standards and needs, given the full specifications of the existing design. 
Even so, constraint programs cannot understand designs in the form of pictures or videos. Nor can they provide compelling visualizations of the generated designs. Moreover, designs generated by constraint programming often satisfy the bare minimum of functionality without considering subtle aspects such as aesthetics and convenience. In fact, it is almost impossible to encode such subtle aspects into the objective function of a constraint program. Therefore, designs generated by constraint satisfaction solely are rarely welcomed by customers. 
This demonstrates a deep drawback in purely symbolic methods like constraint programming -- they cannot engage with the nebulous world of low-level perception and reasoning.
Recent advances in machine learning, in particular deep-generative models, have presented new opportunities to address these challenges. Text-to-image and graph-to-image models provide exciting possibilities for controllable content generation, but their control is not always precise enough to produce designs that meet complex specifications.

For example, in Figure \ref{fig:functionality}, a toaster must be added to the left of the oven and below the sink (already in the image). Additionally, a microwave needs to be added to the right of the oven. In the right panel, the Stable Diffusion (SD) \cite{stable_diff} model (one of the state-of-the-art neural generative models), taking the input of the initial kitchen configuration and the text specifications, simply alters the entire scene, producing results that look pleasing but do not fit the specification. This is emblematic of a deep issue with purely neural algorithms -- thus far, they have failed to grasp the high-level reasoning that symbolic approaches can handle efficiently.
%
Overall, AI-driven automatic design is an emerging field that requires \textit{the integration of neural and symbolic reasoning}. However, such integration is beyond the reach of state-of-the-art models.

\textbf{Our Approach.} We introduce \textbf{Sp}atial \textbf{R}easoning \textbf{In}tegrated \textbf{G}enerator (SPRING) for design production. 
Given an initial indoor scene and user requirements described in propositional logic, the task is to generate a design that satisfies user specifications, looks pleasing and follows common sense. 
%
The essence of SPRING is the \textbf{\textit{embedding of a neural and symbolic integrated spatial reasoning module within a deep generative network}}. 
The spatial reasoning module decides the locations of the objects to be generated in the form of bounding boxes, following an {iterative refinement} approach. 
The bounding boxes are predicted by a sequence-to-sequence neural model and are further filtered by a symbolic constraint reasoning process.
This integrated approach allows us to leverage the advantages of both neural and symbolic methods. 
Here, symbolic constraint satisfaction deals with explicit specifications, such as user requirements, while neural networks handle aesthetics and common sense. 

SPRING consists of three modules. The first perception module based on Detection Transformers (DETR) \cite{detr} extracts existing object positions from input images.

\begin{rev}[1]
It is followed by the spatial reasoning module, which uses a neural and symbolic integrated approach to generate the bounding boxes representing the locations of the objects to be generated. We call a collection of all these bounding boxes a \textit{layout}.
Two factors need to be taken into consideration: first, the satisfaction of explicit spatial specifications, and second, the implicit preferences regarding the naturalness of locations, object sizes,  and aesthetics. 
Our neural and symbolic integrated approach first trains a Recursive Neural Network (RNN) that defines an \textit{implicit preference distribution} over the layouts. The RNN is trained to re-place objects from online images, where these objects are removed using inpainting algorithms. 
In the implicit preference distribution, layouts that meet implicit preferences have high probabilities to be sampled, therefore allowing the spatial reasoning module to capture \textbf{\textit{the implicit spatial knowledge and aesthetics}}. 
We then modify the implicit preference distribution into a backtrack-free distribution, in which layouts that violate explicit user specifications are given 0 probability, and the probabilities of other layouts are scaled up proportionally.
We further define a symbolic reasoning module -- a variant of SampleSearch, initially proposed in \cite{samplesearch} -- to sample layouts from the backtrack-free distribution.
SampleSearch guarantees that layouts violating constraints will never be sampled, thereby ensuring the \textbf{\textit{satisfaction of the explicit user specifications}}.



%
%
%
%
%
%

Finally, the bounding boxes are filled by a Visual Element Generator (VEG) -- currently, diffusion-type models fill this role. 
Diffusion models are trained to generate individual aesthetically pleasing objects from millions of online images. 
In our work, both off-the-shelf Stable Diffusion models and a fine-tuned version are compared for the VEG. 
The VEG is given a small margin around each object to be generated as contextual information. This is to make sure that the generated objects blend into the existing scene aesthetically and seamlessly. 
The three modules of SPRING are illustrated in Figure \ref{fig:rollout}.
\end{rev}

\textbf{How SPRING Advances the State-of-the-art:} 

{\textit{(1) Guaranteed User Requirement Satisfaction via Embedding Symbolic Reasoning into Neural Generative Models}}:
SPRING guarantees that the generated designs satisfy user specifications. 
This is beyond the capabilities of the state-of-the-art neural generative models.

\begin{rev}[2]
\textit{(2)} \textit{Aesthetics and Implicit preference satisfaction}: SPRING also learns to satisfy implicit preferences like aesthetics and naturalness from data. In particular, the spatial reasoning module decides the reasonable locations and dimensions of objects to be placed, and the visual element generator produces detailed visual representations of objects which are blended into the image. Both modules learn from emulating the generation of locations and objects from online images, acquiring a sense of aesthetics, naturalness, and the satisfaction of other implicit preferences. 

{\textit{(3) Interpretable Model:}} SPRING is more interpretable than alternate methods. At each step of iterative refinement, the user can trace the probabilities associated with each sub-decision made by the spatial reasoning module. This helps to remedy potential dissatisfaction of implicit preferences by allowing users to identify and adjust the decision-making process to better align with their design expectations.

\end{rev}

{\textit{(4) Zero-shot Transfer to Novel User Specifications:}} The spatial reasoning module is capable of handling novel constraints in a zero-shot manner. When novel user specifications not present in the training set are given at test time, the symbolic reasoning procedure in the spatial reasoning module still blocks the invalid output from the neural networks in the same way as handling familiar constraints, without the need for retraining or fine-tuning.

SPRING was evaluated in the domain of interior design generation, in which the AI must add elements to an interior space like a kitchen or a living room, taking into account existing objects and constraints given by the user.  In this setting, it is vital that the algorithm produces images that satisfy the user's design specification, while also maintaining a high overall image quality.
Accuracy metrics were collected to objectively measure how well SPRING was able to satisfy spatial constraints from the specification. SPRING was able to produce designs that satisfy complex user specifications, while text-based or scene graph-based baseline approaches could not. 
Regarding image quality, spatial naturalness and general aesthetics are both important factors in a good-looking design. 
Synthetic positioning datasets were created to evaluate SPRING's ability to learn implicit spatial constraints, such as toasters should not be put on the floor, TVs may hang on the wall, etc. In this situation, SPRING was able to outperform a baseline made up of a Generative Adversarial Network (GAN) further filtered by a differentiable convex programming layer.
For general aesthetics, automated image quality metrics were collected in the form of FID and IS scores. The images produced by SPRING were on par in quality with the latest image generation tools. 
As additional support for these evaluations, a human study was conducted to assess SPRING for both specification satisfaction and image quality. The results corroborated the effectiveness of SPRING in generating high-quality images that meet user specifications while maintaining the overall aesthetics and spatial naturalness of the design.
Furthermore, SPRING is capable of handling novel user constraints not presented in the training set in the zero-shot learning environment, which was demonstrated by its ability to satisfy complex novel constraints after training was complete.

\textbf{How SPRING Advances Artificial Intelligence in General:}

SPRING's contribution to the broader field of AI is its successful integration of neural and symbolic approaches. It exemplifies a powerful synergy where the adaptability of neural models is enhanced by the precision and reliability of symbolic methods. This integration addresses a fundamental challenge in AI, offering a balanced solution that leverages the strengths of both methodologies while mitigating their weaknesses. For AI researchers and practitioners across various domains, SPRING provides a practical blueprint for developing more robust, adaptable, and intelligent systems capable of handling complex, real-world tasks with enhanced efficiency and reliability.

\begin{rev}[20]
We chose the interior design domain as a motivating application to spur community interest and highlight the necessity of integrating symbolic reasoning with neural generative models. This strategic focus demonstrates the importance of this integration and enables future work in many other domains. To demonstrate this general utility and extensibility in a small way, as a supplement to our experiments in interior design generation, we also present a trained SPRING instance that produces images of animals in an outdoor setting (see Section \ref{sec:extens}). While this work will largely focus on the interior design domain, SPRING can be applied to constrained image generation more broadly, and future work should explore this integration even further.

This approach echoes human psychology, specifically the theories of dual-process cognition as championed by researchers like Kahneman \cite{slowfast_book}. Within SPRING, two subsystems work together to accomplish a larger cognitive task, with one making snap judgments based on learning and data (the neural component) and one moderating those choices with slower, more reasoned logic (the symbolic reasoning component). The integrated approach shows how AI can benefit from the human cognition process. 
\end{rev}


\FloatBarrier
\section{Problem Definition}

\subsection{Design Production}

Design generation is the problem of producing designs in the form of diagrams, environments, or images. The example chiefly explored here is interior design, in which the image is an interior in a home, such as a kitchen or living room, and the design is specified as a set of objects to add which have positional relationships to each other and objects already in the image. The problem can be defined as:

\begin{problem}
\textit{(Design Production)}: \textbf{Given:} let B be a background image that contains initial objects W, and D be a design specification with new objects O and positional constraints C represented in the propositional design language defined below. C may reference objects in both O and W. T is the set of natural images. 
\\\textbf{Find} a scene image S, with B as a background, containing objects defined by O, such that all of C is satisfied and S is realistic; i.e., it is visually close to the images in T.
\end{problem}

\subsection{Propositional Design Language}
\label{sec:design_lang}
 
\begin{table}[p]
\centering
{\fontsize{11pt}{14pt}\selectfont
\begin{tabular}{l|l}
\hline
\textbf{Relation} & \textbf{Truth condition} \\ \hline
$\textit{above}(o_1, o_2, c)$ & top side of $o_1$ is at least $c$ units above top side of $o_2$.\\ 
$\textit{cabove}(o_1, o_2)$ & bottom side of $o_1$ is at least $c$ units above top side of $o_2$.\\ 
$\textit{above\_value}(o_1, c)$ & top side of $o_1$ has a y-value less than $c$.\\ 
$\textit{below}(o_1, o_2, c)$ & top side of $o_1$ is at least $c$ units below top side of $o_2$.\\ 
$\textit{cbelow}(o_1, o_2)$ & top side of $o_1$ is at least $c$ units below bottom side of $o_2$.\\ 
$\textit{below\_value}(o_1, c)$ & top side of $o_1$ has a y-value greater than $c$.\\ 
$\textit{left}(o_1, o_2, c)$ & left side of $o_1$ is at least $c$ units left of left side of $o_2$.\\ 
$\textit{cleft}(o_1, o_2, c)$ & right side of $o_1$ is at least $c$ units left of left side of $o_2$.\\ 
$\textit{left\_value}(o_1, c)$ & left side of $o_1$ has an x-value less than $c$.\\ 
$\textit{right}(o_1, o_2, c)$ & left side of $o_1$ is at least $c$ units right of left side of $o_2$.\\ 
$\textit{cright}(o_1, o_2)$ & left side of $o_1$ is at least $c$ units right of right side of $o_2$.\\ 
$\textit{right\_value}(o_1, c)$ & left side of $o_1$ has an x-value greater than $c$.\\ 
$\textit{narrower}(o_1, o_2, c)$ & $o_1$ is at least $c$ units narrower than object $o_2$.\\ 
$\textit{narrower\_value}(o_1, c)$ & $o_1$ has a width less than $c$.\\ 
$\textit{shorter}(o_1, o_2, c)$ & $o_1$ is at least $c$ units shorter than object $o_2$.\\ 
$\textit{shorter\_value}(o_1, c)$ & $o_1$ has a height less than $c$.\\ 
$\textit{taller}(o_1, o_2, c)$ & $o_1$ is at least $c$ units taller than object $o_2$.\\ 
$\textit{taller\_value}(o_1, c)$ & $o_1$ has a height greater than $c$.\\ 
$\textit{wider}(o_1, o_2, c)$ & $o_1$ is at least $c$ units wider than object $o_2$.\\ 
$\textit{wider\_value}(o_1, c)$ & $o_1$ has a width greater than $c$.\\ 
$\textit{heq}(o_1, o_2)$ & $o_1$ and object $o_2$ have the same height.\\ 
$\textit{heq\_value}(o_1, c)$ & $o_1$ has a height equal to $c$.\\ 
$\textit{weq}(o_1, o_2)$ & $o_1$ and object $o_2$ have the same width.\\ 
$\textit{weq\_value}(o_1, c)$ & $o_1$ has a width equal to $c$.\\ 
$\textit{xeq}(o_1, o_2)$ & left side of $o_1$ is in line with left side of $o_2$.\\ 
$\textit{xeq\_value}(o_1, c)$ & left side of $o_1$ has an x-value equal to $c$.\\ 
$\textit{yeq}(o_1, o_2)$ & top side of $o_1$ is in line with top side of $o_2$.\\ 
$\textit{yeq\_value}(o_1, c)$ & top side of $o_1$ has a y-value equal to $c$.\\ 
\hline
\end{tabular}
}
\caption{Spatial relations and their truth conditions. Note that integer literals (shown here as $c$) are in units of per-thousanths of the background image's width and height.}
\vspace{-20pt}
\label{tab:spatial-relations}
\end{table}

\noindent\textbf{Symbols and Constants}. 
Our design language uses propositional logic. 
In this language, $o_1, o_2, \ldots, o_N$ denote objects. 
These objects represent furniture or other objects that are either in the background image or need to be added. 
The constants used in the design language are integers and text strings. 
The integers are often used to denote the spatial distances between objects, while text strings define the properties and types of the objects.

\medskip

\noindent\textbf{Properties and Types.}
\texttt{Property} is a predicate that evaluates to true if and only if an object has a given property. 
For example, $\texttt{property}(o_1, ``blue'')$ is true if and only if object $o_1$ has the property $``blue''$ (i.e., its color is blue). 
A special property is called \texttt{type}. This defines the type of object being reasoned over from a set of known types. For example, $\texttt{type}(o_1, ``microwave'')$ evaluates to true if and only if $o_1$ has the type $``microwave''$. 
Combining with the previous example, $\texttt{type}(o_1, ``microwave'') \wedge \texttt{property}(o_1, ``blue'')$ means object $o_1$ is a blue microwave. 
Our language recognizes the following types: chair, couch, potted plant, bed, mirror, dining table, window, desk, toilet, door, TV, microwave, oven, toaster, sink, refrigerator, and blender.

\medskip
\noindent\textbf{Relations}. 
Relations are used to model spatial constraints between objects, providing a way to describe the relative position, size, and alignment of objects within a design. These relations evaluate to true if the spatial relationship is upheld between the objects included, and false otherwise. Our grammar includes various types of predicates to express these relationships (a complete description is available in Table \ref{tab:spatial-relations}).

\begin{enumerate}
    \item \textbf{Spatial relationships with constant offsets:} This group of predicates defines spatial relationships between two objects with a constant offset. Predicates like $\texttt{above}$, $\texttt{below}$, $\texttt{right}$, and $\texttt{left}$ describe relative positions between objects in the x and y dimensions. 
    For example, $\texttt{above}(o_1, o_2, k)$ is true if and only if object $o_1$ is above object $o_2$ by at least $k$ vertical units. More precisely, the top side of $o_1$ is at least $k$ vertical units above the top side of $o_2$. 
    In this paper, we define one vertical unit as one-thousandth (per-mille) of the image's height, and similarly, one horizontal unit as one-thousandth of the image's width.
    Sometimes the distance $k$ is 
    omitted. In this case, $\texttt{above}(o_1, o_2)$ is true if and only if $o_1$ is above $o_2$. 
    Additionally, the grammar provides $\texttt{cabove}$, $\texttt{cbelow}$, $\texttt{cright}$, and $\texttt{cleft}$ predicates, which represent complete spatial constraints where the entire bounding box of one object is constrained in one direction away from its counterpart, with no overlap. For example, $\texttt{cabove}(o_1, o_2, k)$ holds if and only if object $o_1$ is completely above object $o_2$ with a constant vertical distance of at least $k$. In other words,  the bottom side of $o_1$ is at least $c$ vertical units above the top side of $o_2$.     

    \item \textbf{Size comparisons between objects:}
    $\texttt{shorter}$, $\texttt{taller}$, $\texttt{narrower}$, and $\texttt{wider}$ predicates define the size constraints between objects. 
    For example, $\texttt{shorter}(o_1, o_2, c)$ evaluates to true if and only if object $o_1$ is at least $c$ vertical units shorter than object $o_2$.
    Again, the argument $c$ can be omitted. 

    \item \textbf{Equality constraints:} These predicates establish equal attribute relationships between objects. For example, $\texttt{xeq}(o_1, o_2)$ holds if and only if objects $o_1$ and $o_2$ share the same x-position (meaning they are vertically aligned). Similarly, $\texttt{yeq}$, $\texttt{weq}$, and $\texttt{heq}$ predicates ensure that two objects share the same y-position, width, or height, respectively. 

    \item \textbf{Constraints with constant values:} This group of predicates sets specific constraints on an object's attribute with a constant value instead of in reference to another object. For example, $\texttt{above\_value}(o, k)$ holds if and only if the y-position of the top-left corner of object $o$ is above (less than) $k$. Other predicates like $\texttt{below\_value}$, $\texttt{right\_value}$, $\texttt{left\_value}$, $\texttt{shorter\_value}$, $\texttt{taller\_value}$, $\texttt{narrower\_value}$, \\$\texttt{wider\_value}$, $\texttt{xeq\_value}$, $\texttt{yeq\_value}$, $\texttt{weq\_value}$, and $\texttt{heq\_value}$ constrain an object based on given values.
\end{enumerate}

\medskip
\noindent\textbf{Complex Relationship}.  
The complex spatial relationship among objects can be defined using 
logic operators ($\wedge$ for “and'', $\vee$ for “or'', $\neg$ for “not'') to connect the set of predicates discussed above. 
For example, ``a cozy brown leather chair is either completely left or completely right of the black and white striped couch.'' can be described as:
\begin{align*}
&\texttt{type}(o_1, ``chair'') \wedge
\texttt{property}(o_1, ``cozy'') \wedge
\texttt{property}(o_1, ``brown'') \wedge \\
&\texttt{type}(o_2, ``couch'') \wedge
\texttt{property}(o_2, ``black\ and\ white\ striped'') \wedge \\
&(\texttt{cleft}(o_1, o_2) \vee
\texttt{cright}(o_1, o_2)).
\end{align*}

This completes our definition of the propositional design language. 
While natural language is usually easier to use than structured languages like ours, it may suffer from ambiguities, which render it difficult to satisfy hard rules. 
Nevertheless, we intend to eventually extend our work to consider natural language as well. 
We leave such effort for future work.

\FloatBarrier
\section{\texorpdfstring
{\underline{Sp}atial \underline{R}easoning \underline{In}tegrated \underline{G}enerator}
{Spatial Reasoning Integrated Generator}}

SPRING is motivated by the desire to smoothly integrate the handling of implicit preferences from data and explicit rules from the user to generate high quality design images. It does this by integrating a symbolic search component to enforce the rules with a neural component for learning implicit preferences, perceiving the background, and drawing the image. Following this, SPRING is composed of a neural perception module, a hybrid spatial reasoning module, and a neural visual element generation module. 
\begin{rev}[21]
Finally, SPRING includes an optional human feedback step, allowing the constraints to be edited based on initial results.
\end{rev}

\subsection{Perception Module}
\begin{rev}
The purpose of the perception module is to extract information about the existing background for the SRM -- which is utilized in iterative refinement both for implicit preference and explicit rule satisfaction. The perception module includes an \textit{object detector} for predicting existing objects and their bounding boxes, and a \textit{scene encoder} for scene-level feature extraction.

\subsubsection{Object Detector for Prior Visual Element Recognition}
The object detector is implemented with a pre-trained DETR50 \cite{detr} detection transformer model trained on the COCO dataset \cite{coco}, but this can be jointly trained with the spatial reasoning module given a new dataset. 
DETR50 detects objects by taking an image as input and outputting a set of vectors, each representing a detected object. These vectors include a categorical vector for class probabilities and a 4-vector for bounding box coordinates. 
It works by using a convolutional neural network (CNN) \cite{convnets1,convnets2} to extract features from the image, which are then processed by a transformer to predict object locations and classes. The CNN encodes the image into a feature map, and the transformer decodes this map into object predictions, refining them through multiple stages.

\subsubsection{Scene Encoder for General Visual Features}
\label{sec:scene_encoder}

A Resnet18 \cite{he2016residual} architecture pre-trained on ImageNet \cite{deng2009imagenet} is used as the scene encoder, which is fine-tuned during training in conjunction with the SRM -- these networks together are fully-differentiable, so can be trained together as one larger network. 
The features it encodes are used in the SRM, and represent general information about the image, capturing details like lighting, texture, and room composition beyond the objects present.
Resnet18 works by using a series of convolutional layers with residual connections to extract hierarchical features from the input image, gradually building up a representation that captures increasingly complex patterns. Following common practice, the final layer of the Resnet is replaced with a fresh (newly-initialized) one to facilitate the fine-tuning. This final layer outputs the 500-sized vector used later by the SRM to initialize its hidden state.

\end{rev}

\FloatBarrier
\subsection{Spatial Reasoning Module (SRM)}

\begin{rev}
The Spatial Reasoning Module (SRM) is responsible for determining object locations within a design. Our SRM combines neural networks and symbolic reasoning to ensure the satisfaction of both explicit constraints and implicit preferences.
Our SRM samples object locations from the modification of an \textit{implicit preference distribution} learned by a Recurrent Neural Network (RNN) to capture implicit preferences.
The modified distribution, which we call the \textit{backtrack-free distribution}, uses symbolic reasoning to zero out the probability of producing object locations that violate explicit constraints. 
In both distributions, object locations are determined by their respective bounding boxes, which include parameters for the $x$ and $y$ coordinates of the upper-left corner, as well as the width and height. Together, these bounding boxes constitute the \textit{layout} (definition from \cite{zhao2019image}). The SRM receives background image information, including positions and attributes of existing objects, from the perception module. It then samples a layout from the backtrack-free distribution.

Both the preference and the backtrack-free distributions define the probability of generating a layout through an \textbf{\textit{iterative refinement}} process, \textbf{which progressively narrows the range of each parameter through a series of decisions.}
Each decision refines the possible values for one parameter, reducing the range until its precise value is found. 
For example, Figure \ref{fig:srm_tree} (left) decides the range of object 2's $x$ coordinate. 
It starts by assuming this coordinate can take any value in the range between 0 and 8.
In one decision, the range shrinks to the left side (0 to 4), or the right side (4 to 8), or the middle value (4). 
Such decisions continue, forming a tree.
The probability of taking decisions at each step is marked in the figure. The probability of setting the coordinate to a value is then the probability of a decision sequence in the tree. 

The \textit{implicit preference distribution} assumes that the decision probability at each step is the output of a recurrent neural network (RNN), which is trained to capture implicit preferences such as aesthetics, naturalness, and utility. 
See Figure \ref{fig:srm_search} for an example.
The RNN learns such implicit preferences by re-generating the locations of objects removed from existing online pictures using inpainting. The details are in Subsection \ref{sec:learning_in_srm}. 


The \textit{backtrack-free distribution} ensures that explicit constraints are respected. \textbf{It modifies the implicit preference distribution by assigning zero probability to decisions that lead to constraint violations.} For instance, in Figure \ref{fig:srm_tree} (right), the backtrack-free distribution avoids paths that violate user specifications while remaining proportional to the implicit preference distribution elsewhere. Notice that some variables become zeroed out, but the others in the same decision step are normalized to remain proportional.

\textbf{A SampleSearch-based \cite{samplesearch} procedure is used to sample from the backtrack-free distribution}, as seen in Figure \ref{fig:srm_search}. 
SampleSearch is a systematic search procedure including backtracking, normalization, and pruning, taking as input the single-step probabilities generated by the RNN. Its interplay between the neural RNN-component and the symbolic search component ensures that the produced layout is a random sample from the backtrack-free distribution, respecting both the implicit preferences and explicit constraints.

\begin{figure*}[tbhp]
    \centering
    \includegraphics[width=\linewidth]{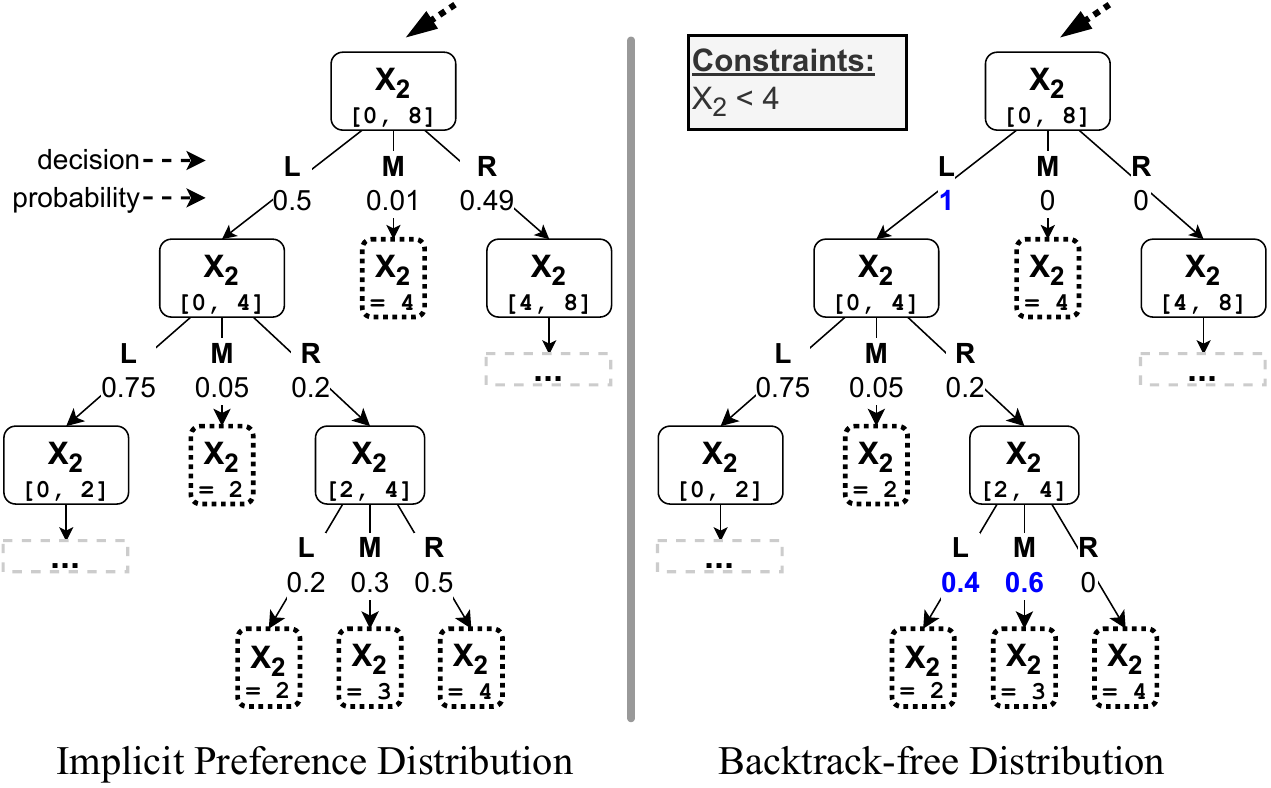}
    \vspace{-10pt}
    \caption{Example problem demonstrating the implicit preference and the backtrack-free distribution of one variable $x_2$. In the root node, $x_2$ can take values between 0 and 8. Such a range is \textit{iteratively refined} in each step (three decisions: taking the left side (L), middle (M), right side (R)) with different probabilities, forming a tree. \textbf{(Left)} The implicit preference distribution defines how the next decision should be sampled from the current distribution. The probability is generated by the RNN. \textbf{(Right)} The backtrack-free distribution, which is proportional to the implicit preference distribution for all decisions that adhere to constraints, but have 0 probability when any are violated. Notice how distributions with new zeros in the first and third levels are renormalized, but still proportional.}
    \label{fig:srm_tree}
\end{figure*}

\begin{figure*}[tbhp]
    \centering
    \includegraphics[width=\linewidth]{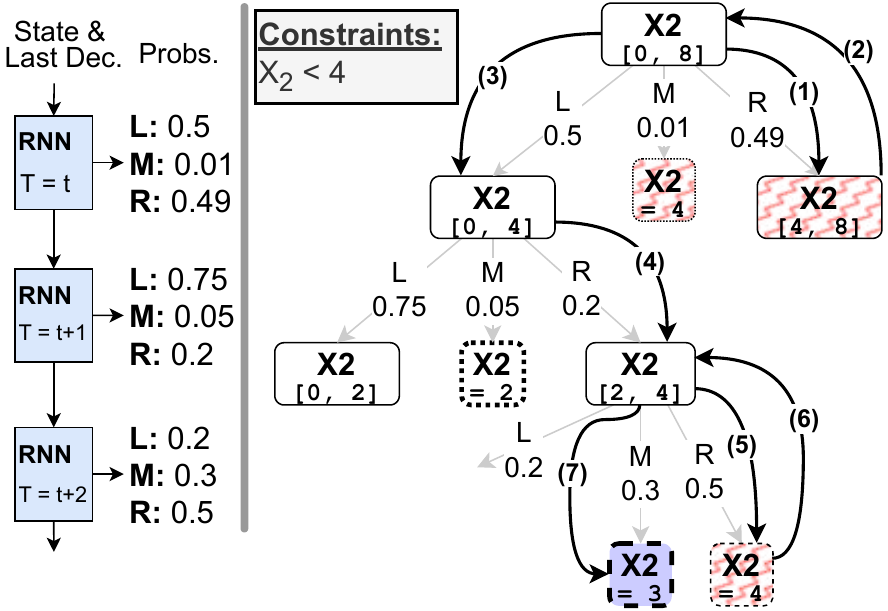}
    \vspace{-10pt}
    \caption{Example from Figure \ref{fig:srm_tree} with the SampleSearch procedure overlaid. \textbf{(1-7)} denotes the 7 steps in the search. \textbf{(1)} Sample ``R'' from distribution. \textbf{(2)} No element of the range is less than 4, so the current node is pruned. Backtrack and normalize distribution. \textbf{(3)} Sample ``L''. \textbf{(4)} Sample ``R''. \textbf{(5)} Sample ``R''. \textbf{(6)} A leaf node has been reached, but it does not satisfy constraints. Backtrack and normalize. \textbf{(7)} Sample ``M''. This leaf node satisfies all the constraints and is returned.}
    \label{fig:srm_search}
\end{figure*}

\subsubsection{Decisions in Iterative Refinement}

In SPRING, iterative refinement is broken down into three types of decisions: left side, right side, and middle. These decisions systematically reduce the range of possible values for each parameter. A left side ($L$) decision narrows the range to the lower half, setting the maximum value to the midpoint. A right side ($R$) decision narrows the range to the upper half, setting the minimum value to the midpoint. A middle ($M$) decision selects the midpoint as the final value, making the range deterministic.

The problem is then formulated as a sequence of these decisions, starting at the root of a tree where all ranges are set to initial values (in our experiments, $[0, 1000]$ were used), and concluding at a leaf where all ranges are a single integer value. To streamline the problem going forward, we assume a set order of decisions (see  Section \ref{sec:ordered_simplification}), where a parameter is reduced to a single value before the next begins refinement.

\subsubsection{The Implicit Preference \&
Backtrack-free Distributions}
\label{sec:implicit_backtrack}
Two probability distributions are important to understanding the SRM: the implicit preference distribution and the backtrack-free distribution. \textbf{The implicit preference distribution} acts as a proposal distribution, representing implicit user preferences at various decision points. It is defined over decisions rather than directly over spatial parameters. The implicit preference distribution over all decisions on all spatial parameters can be defined as follows:
\[ 
Q(\Delta) = \prod_{i=1}^{n} \prod_{k=1}^{K_i} Q_i^k(\Delta_i^k \mid \Delta_{<i}, \Delta_i^{<k}, a, b)
\]


In more detail, the set $\Delta$ includes all decisions for all spatial parameters. The set $\Delta_i$ consists of all decisions refining a parameter $i$, while $\Delta_i^k$ is the single decision for parameter $i$ at decision step $k$. $\Delta_i^k$ belongs to the set $\{L, M, R\}$ -- the domain of possible decisions, left, middle, and right. $K_i$ represents the number of steps needed to reduce the range of parameter $i$ to a single integer. $\Delta_{<i}$ is shorthand for $\Delta_{0}, \ldots, \Delta_{i-1}$, which includes the set of decisions for all previously refined parameters. Similarly, $\Delta_i^{<k}$ refers to the set of decisions for the current parameter $i$ but all previous decision steps. $Q(\Delta)$ represents the implicit preference distribution over all decisions. $Q_i^k$ is the component of the distribution for the single decision $\Delta_i^k$ at decision step $k$ for parameter $i$. The initial integer range for all spatial parameters is set to a constant $[a, b]$, which in this case is $[0, 1000]$.

\textbf{The backtrack-free distribution} is constructed from the implicit preference distribution, maintaining proportionality for decisions that can lead to a constraint-satisfying result, while assigning a probability of zero to decisions that always violate constraints. This ensures that configurations of decisions drawn from the backtrack-free distribution closely follow the implicit preference distribution but will never violate explicit constraints, as any decision leading to a violation is given zero probability. The backtrack-free derivation $Q'$ of the implicit preference distribution $Q$ can be constructed as follows:

\begin{multline}
Q_i^{'k}(\Delta_i^k \mid \Delta_{<i}, \Delta_i^{<k}, a, b) =\\
\begin{cases}
\alpha Q_i^k(\Delta_i^k \mid \Delta_{<i}, \Delta_i^{<k}, a, b) & \text{if } \Delta_{<i}, \Delta_i^{<k} \text{ satisfies constraints} \\
0 & \text{otherwise}
\end{cases}
\end{multline}

We use $\alpha$ to normalize after eliminating decisions that lead to constraint violations, maintaining proportionality. Specifically, $\alpha$ is computed as:

\[
\alpha = \frac{1}{1 - \sum_{\Delta_i^k \in B^{\Delta_{i-1}}_i} Q_i^k(\Delta_i^k \mid \Delta_{<i}, \Delta_i^{<k}, a, b)}
\]

where $B^{\Delta_{i-1}}_i$ is the set of decisions for parameter $i$ at step $k$ that lead to constraint violations, and $Q_i^k(\Delta_i^k \mid \Delta_{<i}, \Delta_i^{<k}, a, b)$ is the probability of making decision $\Delta_i^k$ given the previous decisions and initial range. 

For example, consider a decision where we are refining the width parameter for object 2. The current range is $[0, 250]$ and the probability distribution over decision values $\{L, M, R\}$ is $\{0.2, 0.3, 0.5\}$. However, the middle decision will violate constraints. Therefore, in the backtrack-free derivation, the probability $\Delta_i^k = M$ is set to 0.0. $\alpha$ resolves to $\frac{1}{1 - 0.3} = \frac{1}{0.7} \approx 1.429$, and the final backtrack-free derivation is $\{0.286, 0.0, 0.714\}$. This results in a 71\% chance of choosing $R$ and refining to a range of $[125, 250]$, and a 29\% chance of choosing $L$ and refining to a range of $[0, 125]$. In either case, $M$ is never sampled.

\subsubsection{Parameter Sampling}

The SRM's sampling method is a variant of SampleSearch \cite{samplesearch}, taking the implicit preference distribution over single-step decisions from the RNN as input. 
SampleSearch expands the search tree 
in a depth-first order. It maintains a range table at each search node, listing the current ranges of each coordinate.
At each node, it samples a decision (L, R or M) from the current probability table at the node and descends into the branch. 
%
It continues sampling decisions and descending if no constraint violations are encountered. 
When it descends into a node violating constraints, it backtracks and updates the probability tables of its parents.

\textbf{Range table.} The range table holds the current ranges of possible values for each spatial coordinate. It tracks the minimum and maximum values for each of the four spatial variables of every object. Thus, for $|O|$ objects, the range table maintains $|O| \times 4 \times 2$ values. A range table is considered \textit{deterministic} if all of its parameters have a single integer value in their range, and \textit{non-deterministic} otherwise.

\textbf{SampleSearch Algorithm.} SPRING's sampling algorithm is a search that begins at the root of the iterative refinement tree, where all parameters have the range $[0, 1000]$. The first parameter is selected for refinement. At the first decision, three choices are available: left side, right side, and middle. The RNN, having been trained (see Section \ref{sec:learning_in_srm}), can produce the implicit preference distribution over single decisions. The RNN outputs a softmax distribution on all 3 decisions ($Q_i^{k}$). The choice for the current action is sampled as a multinomial distribution $\delta_i^k \thicksim Q_i^{k}(\Delta_i^k | \Delta_{<i}, \Delta_i^{<k}, a, b)$, deciding the next node of the search. Each node will have its own configuration of the range table, which is created and updated accordingly. This process continues until the current parameter becomes deterministic, then the next parameter in the ordering is chosen, and sampling continues on that parameter's decisions.

In the absence of any constraints, SampleSearch will continue until reaching a deterministic node representing a full layout. 
It will return this layout as a sample.
Along the way, each single-step decision is sampled from the implicit preference distribution provided by the RNN. Therefore, the sampled layout meets implicit preferences such as reasonable spatial arrangements and aesthetics. 


If SampleSearch reaches a leaf node violating constraints, it backtracks to its parent node and modifies its probability table --  resetting the probability of choosing the child node to 0 and normalizing the probabilities of taking other decisions. It then samples another decision from the parent node and continues the search until finding a satisfying node. If an internal node has all of its children set to 0 probability, then that internal node is also deemed as violating constraints. In this case, we backtrack to its parent node, and its probability becomes 0. If the probabilities of taking all decisions at the root node are 0, we must conclude that there is no layout that satisfies constraints. In this case, we return ``no solution'' statement to the user along with the search tree as a trace of reasoning.

This search also includes a pruning step that trims branches which certainly lead to constraint violation. This pruning, as well as further details on checking nodes for constraint violation, can be found in \ref{sec:con_check_prune}.

\begin{proposition}
The layouts sampled from SampleSearch procedure are from the backtrack-free distribution defined in Subsection \ref{sec:implicit_backtrack}.
\label{prop:samplesearch}
\end{proposition}

The proof of Proposition \ref{prop:samplesearch} follows from the proof of Theorem 1 in \cite{samplesearch}.




\subsubsection{Learning Implicit Preference}
\label{sec:learning_in_srm}

In SPRING's spatial reasoning module, the RNN’s role is to model the implicit preference distribution for each decision based on prior decisions, the input image, and the object class being added (e.g., chair, couch, table). This implicit preference distribution captures implicit qualities like aesthetics and naturalness, which are best learned from data. Deep learning methods, such as RNNs, excel at extracting this type of nuanced knowledge from datasets.

\textbf{RNN architecture.} Our RNN neural component is implemented using gated recurrent units (GRUs) \cite{gru}. The background image is encoded by an encoder network as the initial hidden state of the RNN. Starting from this state, the SRM decides the bounding box of each object to be generated sequentially. Moreover, the locations of the bounding boxes are further filtered by a symbolic reasoning module to ensure constraint satisfaction. Specifically, the pipeline is made up of 3 leaky-ReLU-activated gated recurrent units followed by a dense layer, and a Softmax activated output layer. Each GRU inputs and outputs a 4-size vector. GRUs use a hidden state to preserve information across time, and manage that information using internal (learned) reset and update gates. This hidden state vector (size 500) is initialized from a vector produced by the scene encoder within the perception module (see Section \ref{sec:scene_encoder}). The hidden state is also updated by a simple variable encoder, which produces a vector for each positional variable of each object to let the GRUs know that a new object or variable has started, and to encode the object class and variable type. This encoding is added (summed) to the state vector directly.

The output of the GRU is discarded, and the hidden state is instead passed to the dense layer. This is a common practice with GRUs to encode the temporal aspect of the problem within the hidden state and then pass that state to a dense layer. The dense layer has 512 internal units and 4 output units. The basic unit of this RNN output is a \textit{decision token} -- a size 4 vector representing a score distribution of three possible decisions and one special start token. The input at each timestep is the previous timestep's decision  as a one-hot vector. The network outputs at each step a softmax vector to assign the probability to each token. 

\textbf{Training of the RNN.} The training of SRM uses a dataset containing background images with a bounding box for each object to be generated. Each object's location is defined by the upper left point, as well as the width and height of the bounding box. These variables are converted to decision strings as described in the previous sections. The training process is a supervised procedure -- it trains the spatial reasoning module to generate bounding boxes (in the form of decision strings) which match those in the training dataset. This procedure teaches the neural network implicit spatial knowledge including the relative sizes of objects and their common spatial relationships (toasters go on the counter, televisions might be hung on the wall, etc). 

The training of RNN is accomplished in-part through \textit{teacher forcing} \cite{teacher_forcing}. In teacher forcing, the ground truth decision token from the previous step drawn from the training set is used as the input to train the RNN to predict the decision of the current step. Teacher forcing can drastically speed up RNN training, but can also result in lower resilience to uncommon sequences. For this reason, a mixture of strategies -- with and without teacher forcing -- is used to train the model. 50\% of the time the input of the RNN is set by teacher forcing and 50\% set by the predictions of the RNN itself. We also randomize the order of the objects in the decision string during training. 

\FloatBarrier

\subsubsection{The SRM Satisfies Explicit Constraints}
Because the SRM's sampling method blocks all samples that will lead to certain constraint violation (by setting them to probability 0), no decision will be finalized which violates constraints. The SRM ensures that all possible configurations are searched until a satisfying configuration is found, prioritizing those with high preference through the sampling process. If a configuration violates constraints, it is backtracked, and its probability is reset to 0, allowing the search to continue with normalized probabilities for remaining options. This exhaustive process guarantees that any final decision adheres to the explicit constraints. If no configuration is found after this exhaustive search, the SRM returns a ``no solution'' statement to the user, along with a trace of the reasoning, ensuring transparency and adherence to specified constraints.

\subsubsection{The SRM Respects Implicit Preferences}
Because the RNN has learned from a dataset that complies with implicit preferences (such as aesthetics, naturalness, and utility), the implicit preference distribution it generates takes these factors into account. It captures the nuanced and often fuzzy knowledge of how objects should be arranged to be visually appealing and functional. The backtrack-free distribution maintains the proportions from this implicit preference distribution, ensuring that the SRM's sampling method, which is equivalent to the backtrack-free distribution, optimizes decisions for these implicit preferences. Consequently, the decisions made during the layout formation process are not only compliant with explicit constraints but also optimized for implicit qualities, resulting in designs that are aesthetic and practical.

\subsubsection{Ordered Simplification}
\label{sec:ordered_simplification}
We use a set (arbitrary) ordering of how spatial parameters are considered. Example, object 1 x-value, then object 1 y-value, etc. This reduces load on the RNN component, as for what it needs to learn.

\subsubsection{Incorporating Existing Objects}
During design generation, existing objects perceived in the background image may influence the decisions made by the SRM. Not only should the user be able to set positional constraints that reference existing objects, but these objects must be encoded in the neural component of the SRM as well. This is because the neural component is an RNN with a state that is stored throughout the generation process, and encoding what objects are currently in the room will give it vital context for future objects. This is handled by making no distinction between objects that already exist in the background -- those found by the perception module -- and objects that need to be generated in the scope of the SRM. Each of the existing objects is constrained to their already-known parameter values (using constraints like $\texttt{xeq}$, $\texttt{yeq}$, etc) but they are still processed through the SRM to update the hidden state. As the constraints are set to a single value, every decision is forced down a particular path. These are processed through the SRM first before new objects.
\end{rev}

\subsection{Visual Element Generator (VEG)}
The  \textbf{\textit{visual element generator (VEG)}} takes as input the background, the prompt, and location of each object in the form of a bounding box, and outputs an image patch for each bounding box which contains the requested visual element that can be merged seamlessly into the background image in a process conventionally known as \textit{inpainting}. The result is a completed scene adhering to the design with all positional constraints guaranteed. The visual element generator is sequentially called on each bounding box created by the spatial reasoning module.

\begin{rev}[7]
Our SPRING framework offers a flexible structure where various neural generative models can be used for the VEG.
In the experiments, we utilized both off-the-shelf Stable Diffusion-based neural generative models and fine-tuned versions of these models for indoor scene generation.
In each case, the bounding box generated by the SRM and the prompt information are provided as input to the model, and the output is an image patch that can be integrated into the background.
Since the Stable Diffusion-based VEG is trained on a vast collection of online images, it can produce detailed and visually appealing image patches of individual objects, meeting the implicit preferences and aesthetic requirements.

\subsubsection{Generative Models based on Stable Diffusion}

\begin{figure*}[p]
    \centering
    \includegraphics[width=\linewidth]{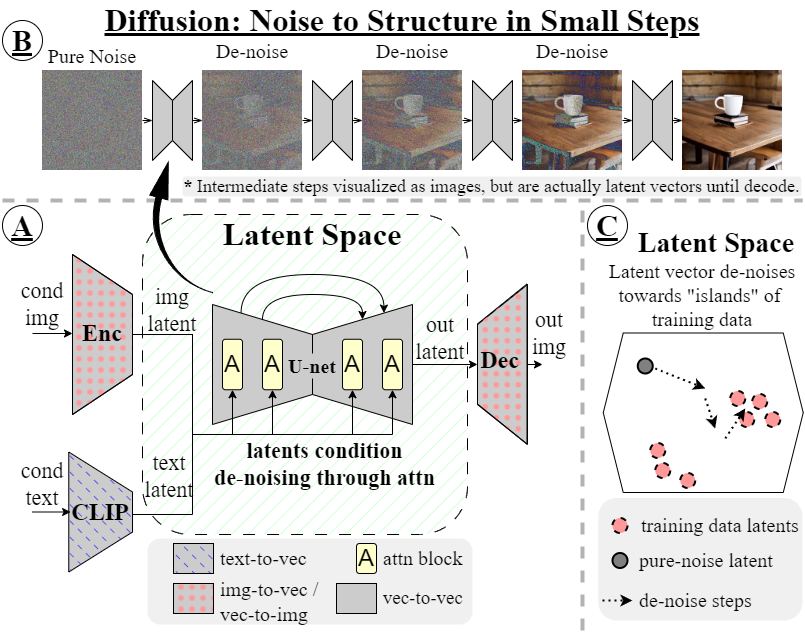}
    \vspace{-10pt}
    \caption{Typical latent diffusion approach, as utilized by our Visual Element Generator. \textbf{(A)} An image can be generated conditioned on text and image inputs. These inputs are transformed to the latent space by CLIP \cite{clip} and an autoencoder, respectively. \textbf{(B)} Diffusion starts with a pure noise vector and removes noise in small steps to form an image. This occurs entirely in the latent space by the U-net \cite{unet}. \textbf{(C)} In the latent space, each de-noising step is a soft adjustment towards a space more familiar to the U-net. These ``islands'' are formed during training, and map to images that look good, and fit the conditioning data.  \vspace{-8pt}}
    \label{fig:lat_diff_diagram}
\end{figure*}

Stable Diffusion from latent spaces follows a series of auto-denoising steps, which generates an image from a vector in the semantic space. This process is visualized in Figure \ref{fig:lat_diff_diagram}. Starting with pure Gaussian noise, a latent vector is de-noised by a central network in small steps. This guides the latent vector towards a comprehensible structure as defined by the training data -- the final vector, when decoded, produces a good-looking image. By conditioning this central network on latent representations of text or input images, the output image will be pushed towards semantically aligning with both. In other words, if conditioned on the text ``a green chair'', the output latent vector will become progressively more similar to the vector embedding of a green chair.
The networks of Stable Diffusion are composed of a U-Net \cite{unet}, an image autoencoder, and a text encoder. Its function is to generate 512x512 images conditioned on a same-size context image and a string of natural language text. 



\textbf{Autoencoder / Pixel-to-Latent Space.} A context image like the background can be input through the autoencoder, which encodes the image into a latent space representation by compressing the image into a lower-dimensional space that retains the essential features of the image. The autoencoder is structured with an encoder that compresses the input image into a latent space and a decoder that reconstructs the image from this latent representation. The encoder consists of residual convolution blocks and self attention blocks, slowly shrinking the dimensionality of the data. The decoder mirrors the encoder’s structure in reverse, transforming the latent representation back into a detailed image.

\textbf{Text Encoding.} Text is input through the Contrastive Language–Image Pre-training (CLIP) \cite{clip} encoder.  CLIP is trained to align the latent representations of text and images – for example, it might learn that the string ``A brown desk’’ should have a very similar latent vector to an image of a brown desk. Its contrastive pre-training lets it convert the input text into a series of embeddings that capture the semantic meaning of the text in relation to images. 

\textbf{Reverse Diffusion through U-net.} The reverse process of diffusion is performed by the U-Net. The U-Net features convolutional blocks with an encoder-decoder architecture. The encoder path includes convolutional layers and downsampling to capture high-level features, while the decoder path uses upsampling and convolutional layers to reconstruct the image. Skip connections link corresponding encoder and decoder layers, retaining latent spatial information. Attention mechanisms between convolutional blocks enhance focus on relevant image parts, improving output quality.

\textbf{Latent Diffusion.} These parts work together by using the text and image embeddings as conditions for the U-Net during the reverse diffusion process, which iteratively refines a noisy latent representation into a structured latent that, when decoded, produces a coherent, high-quality output image. Using cross-attention, these text and image vectors directly interact with the image's noisy latent vector at each diffusion step.

\textbf{Inpainting with Mask.} In the case of inpainting, a mask can be applied by specifying the area of the image to be altered, and the model fills in the masked region while considering the surrounding context and the provided text prompt. This is applied at each step of the diffusion, so the contextual part remains unchanged and available.
The use of region-limiting bounding boxes in the layout combined with the masking process and the iterative application of objects is the reason SPRING does so little damage to its background, which cannot be said of other methods.

\subsubsection{Using a Pre-trained Model in the VEG}
\textbf{Preparing the Image.} To prepare the image, we build the model with half-precision floating-point weights, and set the number of inference steps to 80. A mask is constructed from the bounding box, where pixels inside the box are set to 1, and all others to 0. Given the size constraint, a 512x512 crop is found around the object's bounding box, ideally centered on the box. If a center crop would exceed the border of the image, it is adjusted by minimally shifting the crop box back within the image boundaries, ensuring no part of the crop lies outside the image. This method preserves as much background context as possible without compromising local image quality. The crop is then applied to both the background image and the mask.

\textbf{Preparing the Prompt.} The prompt is constructed programmatically from the object's type and any properties defined in the constraint language using the $\texttt{type()}$ and $\texttt{property()}$ predicates. For example, ``A [Object type]. [Property 1]... [Property n]'' Alternatively, the user can directly enter a prompt for each object.

\textbf{Inpainting the Object.} The prompt, cropped background, and cropped mask are fed into the inpaint model, and the resultant image is saved. This image represents only the cropped area, so it must be reinserted into the larger image. Because inpainting preserves all outside of the bounding box, and the cropped area exceeds the dimensions of the bounding box, the entire area fits seamlessly into the original background.

\textbf{Fully Pre-trained Models Tested in VEG.} Several of the VEG models tested in this work follow this approach as completely pre-trained models. We evaluate the base Stable Diffusion models with versions 1.2 and 1.4. Additionally, we try a publicly available model pre-trained on Stable Diffusion V1.2 \cite{runwayml-inpaint}, which also included additional inpainting-specific training (with mask images for each object included) drawn from the `` laion-aesthetics v2 5+’’ dataset \cite{laion_aesthetics_v2}.

\subsubsection{Finetuning a Model in the VEG}
SPRING also allows the VEG to be fine-tuned in conjunction with training the SRM, to produce more visually and aesthetically consistent results. The details of this are below.

\textbf{Finetuning Approach.} 
Our finetuning approach is a variant of low-rank finetuning (LoRa) \cite{lora}, which allows changing the behavior of a large diffusion model without succumbing to catastrophic forgetting \cite{cata1,cata2} or requiring many computational resources. This method adds trainable parameters to key locations within the network, particularly around the previously-mentioned self-attention and cross-attention modules inside the U-net. At a high level, the parameters of the original model are frozen so they cannot be changed. Two smaller matrices are created by decomposing the original large parameter matrix. These matrices have a much smaller rank than the original, significantly reducing the number of parameters. Their dot product approximates the original parameter matrix. The frozen original parameters are combined with the adapted low-rank updates, enabling effective finetuning with far fewer parameters. 

Training proceeds as it normally does for Stable Diffusion, albeit only with the LoRA parameters trainable. This involves a supervised approach where the mean squared error loss is calculated between the known noise and the predicted noise. The AdamW \cite{adamw} optimizer is then used to minimize this loss over many training steps, ensuring efficient and effective fine-tuning.

\textbf{Finetuned Models Tested in VEG.} One of the VEG models we tested was produced in this way. A model was initialized from Stable Diffusion v1.4. Low-rank (rank of 4) parameters were instantiated around the U-net. CLIP and the autoencoder were not adjusted. The dataset utilized was the same COCO subset used to train the SRM. Text prompts were produced simply by listing the categories of object documented as being in the image. Random image flips were introduced to supplement the existing data. A batch size of 4 and a constant learning rate of 0.0004 was used. Training proceeded for 8840 timesteps. After finetuning, execution proceeds in the same fashion as a pretrained model.

\subsubsection{Impact on Aesthetics}
The purpose of utilizing massive pretrained models, potentially finetuning them, and integrating them into an iterative application process is to serve the needs of aesthetic quality in design. The SRM and the layout it produces  are sufficient for producing locations that follow user constraints and are natural. This has an effect on aesthetic quality, because naturally positioned objects will be more aesthetically pleasing than unnaturally positioned ones. However, none of that has any meaning unless the images of the objects produced from the layout are not themselves aesthetically pleasing. 

Models trained on massive image datasets, such as Stable Diffusion, learn from a vast array of visual data. This extensive training allows the models to capture a wide range of visual features and styles, enhancing their ability to generate high-quality, aesthetically pleasing images. The models generalize to the subtleties of color, texture, lighting, and composition, all of which are crucial for creating visually appealing designs. 

Finetuning these pretrained models to the specific task of interior design further enhances their aesthetic quality. Through finetuning, the models can adapt to the unique characteristics and requirements of interior design, such as the typical color schemes, object proportions, and spatial relationships found in living spaces. This specialized training allows the models to generate designs that are not only visually appealing but also contextually appropriate for interior environments. 

Perhaps most importantly, SPRING can utilize the full power of an image generation model on every object by calling it iteratively as a series of much simpler inpaints. This approach allows SPRING to focus the model's capabilities on one object at a time, ensuring high-quality, detailed generation for each element in the scene. By using region-limiting bounding boxes and masks, SPRING minimizes the risk of altering the background or other objects, preserving the overall aesthetic integrity of the scene. This iterative application ensures that each object is generated with maximum attention to detail and quality, resulting in a visually pleasing final design.

\subsubsection{Inpainting Failures}
A challenge of any diffusion model is the rare chance that it will generate the wrong object or no object at all. These {inpainting failures} are possible, although our method reduces them since each call is responsible for only a single object placed in a reasonable location. Correctly painted objects are guaranteed to follow positional constraints. Additionally, finetuning processes can further reduce the likelihood of these failures (see Section \ref{sec:veg_compare_sec}).

\end{rev}

\begin{rev}[22]
\subsection{Optional Human Feedback}

SPRING concludes with a simple and optional human-feedback step, in which the resulting image is presented to the user, along with the SRM-produced layout. If the user finds the output unappealing, they can choose to regenerate the output from any step of the process, including altering object-specific prompts to get new images from the same layout, or recreating the layout and all objects with altered constraints. The following examples illustrate this functionality and the potential for adjustment:

\textbf{The first example} (changing only visual properties of the objects, not locational ones) allows more aesthetic control to the user. For example, if the user would like a couch that was generated to be red instead of brown, they could add the property $\texttt{property}(couch, ``red'')$. The same layout would be kept, but the object inpainting would be regenerated with this new property as a part of the prompt.

\textbf{The second example} (changing the spatial and locational constraints on objects) allows more detailed control over the spatial elements, and allows for the correction of overly-loose constraints. Consider the scenario where object $o_1$ is specified to be wider than $o_2$ via the constraint $\texttt{wider}(o_1, o_2)$. Consider also that both objects are of the same type, and it is common within the SRM's training dataset to see two similarly-sized instances of these objects. The process of iterative refinement may settle on $o_1$ being wider than $o_2$ by only a very small margin (e.g. 1 per-mille). This would satisfy the constraint and fit the SRM's understanding of a good scene, but it likely would not satisfy the user. The optional human-feedback step allows this to be remedied, as the user can specify a tighter constraint -- $\texttt{wider}(o_1, o_2, 100)$ would add an offset, forcing the object to be wider by at least 100 per-mille.

\end{rev}

\FloatBarrier
\section{Related Work}
\textbf{Neuro-Symbolic AI.} Neuro-symbolic methods combine neural networks, which excel at pattern recognition and are adaptable to diverse tasks, with symbolic systems, which are adept at reasoning and explicit knowledge representation. This hybrid approach aims to leverage the strengths of both methodologies to create AI systems that can learn from data and reason with symbolic knowledge. This concept has produced numerous recent breakthroughs \cite{Mitchener2022,wu2023spring,mitchell2021conversational,NEURIPS2018_5e388103,NEURIPS2021_d3e2e8f6,mao2018the,galetic2023flexible,Ferry2023Improving}.


Much of the work into neuro-symbolic methods have focused on their much greater explainability than conventional neural methods, without sacrificing the predictive power of those same networks. This has taken many forms, including producing first-order logic explained predictions \cite{gori2023logic,ciravegna2023logic,barbiero2022entropy,ciravegna2020constraint} and concept-based reasoning \cite{marconato2022glancenets,mao2018the}.

Work has also gone into using the formalisms in symbolic reasoning to make neural networks safer \cite{yang2023safe} and more fair \cite{ferry2022improving,davidson2020making,davidson2022towards}. This is broadly possible because symbolic algorithms often provide useful guarantees on their output that go beyond what purely neural methods can supply.

Still other approaches focus on the potential benefits to generalizability that symbolic approaches bring, allowing the system to deal with noise and nebulous concepts using neural approaches, and abstract reasoning using symbolic ones\cite{pmlr-v155-sun21a,mao2018the,evans2018learning,Jaeger2014ConceptorsAE}.

Methods of integration, desired outcomes, and the symbolic approaches used differ. For example, neuro-argumentative learning \cite{toni2023roadmap,ijcai2021p600,baroni2018handbook,cocarascu2016argumentation,albini2020dax,dejl2021argflow,sukpanichnant2021neural,ayoobi2023sparx,cocarascu2019extracting,cocarascu2020data} integrates neural algorithms and symbolic argumentative reasoning, where the latter uses structured logical arguments to process conflicting information. By combining the two, it leverages the pattern recognition capabilities of neural networks to inform and refine the logical arguments, resulting in a system that can adapt to and often explain new data while maintaining structured reasoning capabilities. This question of integration is a major consideration for all parts of the field \cite{745}.
Our work focuses on the integration of neural networks with constraint reasoning, 
in the domain of design generation. 


\textbf{Fast \& Slow Thinking.} Our work also draws inspiration from the fast and slow thinking model described in works from cognitive psychology. Psychophysiological studies have long supported the concept of dual-process thinking in humans \cite{dualproc_creative1,dualproc_creative2} -- in which more deliberate and more intuitive thinking are the result of separate fundamental processes. System 1 refers to a set of innate and learned automatic processes that are rapid, parallel, and unconscious. System 2 is responsible for abstract and hypothetical thinking and is slower, more deliberate, and more conscious \cite{slowfast_book,evans2003two}. These concepts hold exciting parallels for the world of AI, which are already being seriously explored \cite{lecun2022path,NIPS2017_d8e1344e}. Recent work has explored the idea that inspiration can be drawn from these neurocognitive concepts to improve adaptability, generalizability, common sense, and causal reasoning in AI \cite{booch2021thinking}.

\begin{rev}[8]
Our SPRING model integrates Fast and Slow Thinking by pairing neural networks (System 1 / ``fast'') with symbolic reasoning (System 2 / ``slow''), blending modern deep learning's fluidity with traditional symbolic AI's methodicalness. This duality allows SPRING to generate innovative designs that also adhere to precise constraints, but the key concept of bridging deep learning's adaptability with symbolic AI's reliability is widely pursued in AI. As such, our work takes inspiration from concepts like \textit{fast and slow planning} \cite{fabiano2023fast}, with the notable change that our system does not utilize any sort of meta-cognition AI-subsystem \cite{10.1007/978-3-031-25891-6_38} to regulate between fast and slow thinking modules explicitly. 
\end{rev}

\textbf{Constraint Reasoning \& Machine Learning.} Constraint reasoning's integration with deep learning aims to boost reasoning and optimize performance. Historically, this has included incorporating ``common-sense'' reasoning into natural language models \cite{reasoning_hybrid_commonsense, reasoning_simple, reasoning_towards_nn}, case-based reasoning from expert systems \cite{reasoning_case_based}, and relational reasoning between objects or terms \cite{reasoning_simple_mod, reasoning_gnn_neurosymbolic, reasoning_relational_survey, zambaldi2018relational, dvzeroski1998relational}.

Embedded constraint reasoning in machine learning is a promising idea \cite{borghesi2020improving,NEURIPS2021_49e863b1,lombardi2018boosting,bessiere_inductive_2018,rousseau2002using}, providing a structured way to incorporate domain knowledge and improve problem-solving capabilities. Convex optimization has been successfully employed as a neural net layer \cite{cvx_book, cvxpylayers, cvx_control_policies, cvx_opt_models}, and other constraint reasoning methods have been integrated into neural nets for various applications \cite{bartolini2011neuron,detassis2021teaching, reasoning_embed_dd, bai_et_al:LIPIcs.CP.2021.17, khalil2017learning, chen2021automating,mulamba2023learnopt}. Often, this comes in the form of using data to produce constraint models. Constraint acquisition enables learning of constraint networks from examples \cite{BESSIERE2017315, prestwich2021classifier,tsouros2023learnconstr}. Other research is focused on constraint models embedded into networks. For example, decision diagrams have been successfully integrated into GANs for constrained schedule generation \cite{reasoning_embed_dd}, and similar approaches have excelled in text2SQL generation and similar tasks \cite{JMLR:v23:21-1484}.

\textbf{Automatic Design.} Design domains have presented an exciting challenge for AI developers -- recent progress has been made in architecture design \cite{Yoshimura2019}, chip design \cite{khailany2020accelerating}, and biosystems \cite{volk2020biosystems}. Automatic interior design problems have been tackled in several ways previously \cite{appl_review_interiordes} -- including reinforcement learning \cite{rlayout} and conditioning \cite{Chang_2021_ICCV}. Often, the data generated for this task is an abstract diagram of an interior space \cite{9760968,housegan,nauata2021house}, but this approach is hard to visualize for a user. Other works focus on interior space images \cite{interior_sketch_color}, offering better visual qualities but less structure within the data itself. 

\textbf{Text-to-Image \& Graph-to-Image Methods.} Large diffusion models like DALL-E \cite{dall-e,dall-e2}, GLIDE \cite{glide}, and Stable Diffusion \cite{stable_diff} can inpaint objects into scenes given a natural language prompt -- made possible by large online datasets and the combination of transformer \cite{transformer} and diffusion \cite{diffusion} models. However, they fail on complex instructions, such as constraining the number and spatial relationships of objects. 

Another strong line of research encodes requirements into hidden vectors via graph or recurrent nets \cite{sgengnn,sgengan2,sgenrnn,liu2021learning}, and then conditions image generation on these vectors. This includes sg2im \cite{sg2im}, which encodes positional relationships into scene graphs for image generation. These approaches struggle with complex constraints and those not present in training. Other works \cite{zareian2020learning,dhamo2020semantic,garg2021unconditional} adapt this approach for image-to-image problems, using scene graphs as editable latent representations. 
Also related are layout-to-image approaches \cite{zhao2019image,sylvain2021object} which generate an image from a background and user-supplied layout. Most recently, this has been achieved with ControlNets \cite{zhang2023adding}, which allow efficient tuning to new conditions for large diffusion models. Yet, these all require the positions of all objects to be strictly and completely defined by users (e.g., as sketches) and do not reason about positions or spatial relationships.

\FloatBarrier
\section{Experiments}

Evaluating the performance of SPRING for interior design production involves examining three important aspects of the algorithm: the ability to consistently meet design requirements (i.e., constraints), the naturalness of the Spatial Reasoning Module's spatial predictions, and the realism of the final images produced by SPRING. It is important to emphasize that assessing the quality of generated images is a challenging problem in the field, and determining the naturalness of positions and dimensions, independent of aesthetic quality, is even more difficult.
\begin{rev}[11]
Furthermore, specific implicit preferences like aesthetics and convenience are hard to measure, and the impact of satisfying constraints on the realism, aesthetics, and naturalness of the generated designs is challenging to assess. 
\end{rev}
To tackle these challenges, the first aspect (constraint satisfaction) was evaluated using synthetic data scenarios for precise measurement, while the second and third aspects (quality of images, naturalness of positions and dimensions) were trained on real data. To address the known issues with automatic metrics commonly employed in image generation tasks, a human survey was also conducted.
\begin{rev}[11]
Essentially, we take a holistic approach by demonstrating that we can learn implicit preferences generally, and then showing that the end products are realistic, aesthetic, and natural through FID and IS scores, as well as our human study.
\end{rev}

\begin{rev}[13]
Combining all of these experiments, we seek to answer several key questions:
\begin{itemize}
\item \textbf{Does SPRING produce design scenes that fit the specification better than baselines?}: We evaluate this through design adherence experiments, and through scores obtained in a human study.
\item \textbf{Does SPRING have superior image quality to baselines?}: This is evaluated in the image quality experiments using FID and IS scores, the human study, and a targeted spatial reasoning experiment to assess preference accuracy.
\item \textbf{Is SPRING easily extensible?}: This is checked by demonstration of zero-shot constraint adaptation, as well as by tested and training an instance of SPRING on an outside domain.
\end{itemize}
\end{rev}

\begin{rev}[15]
\subsection{Comparison of VEG Backbone Models}
\label{sec:veg_compare_sec}

\begin{figure}[t]
    \centering
    \includegraphics[width=0.8\linewidth]{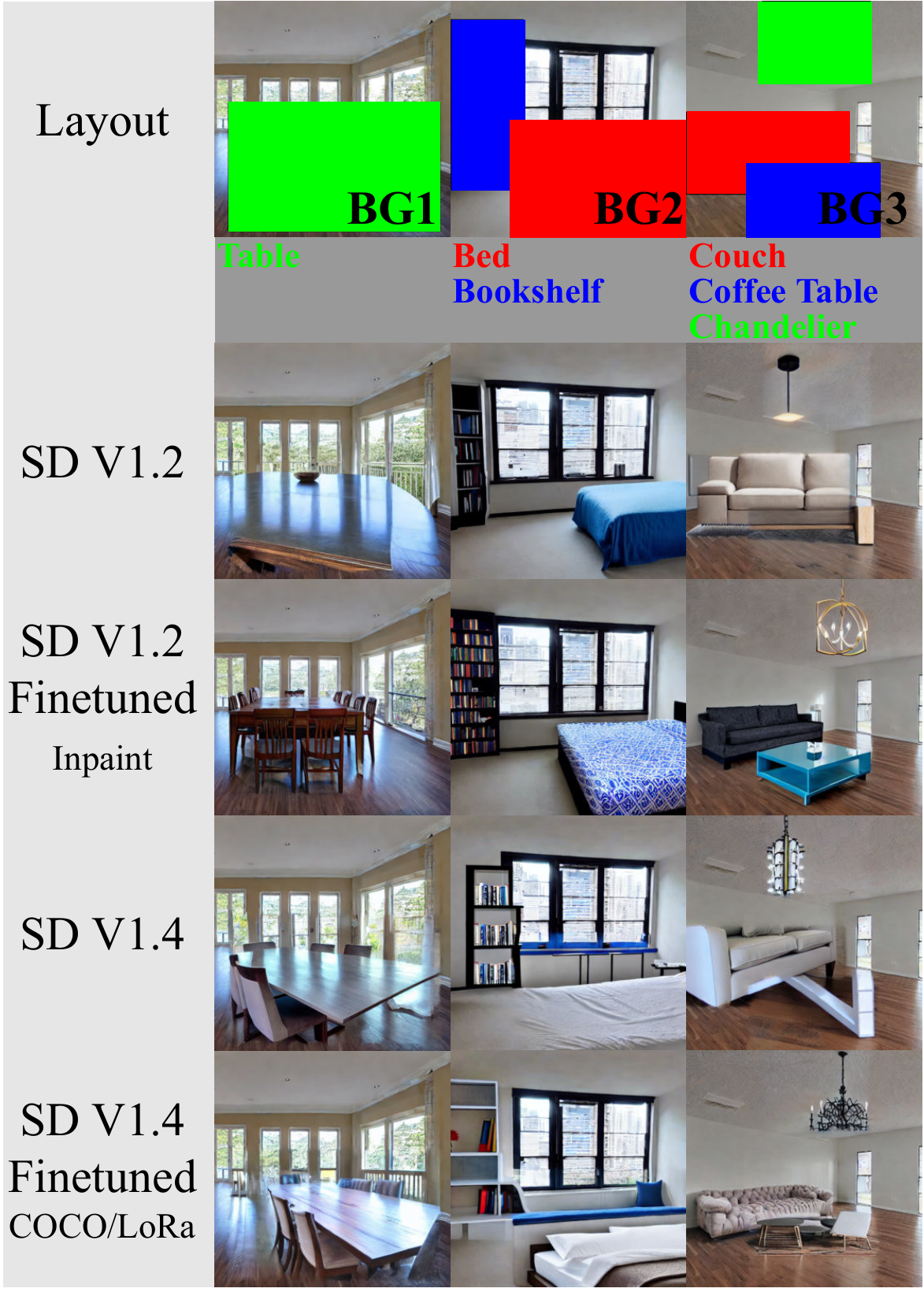}
    \vspace{-8pt}
    \caption{Examples from the Stable Diffusion VEG comparison experiment showing three different backgrounds (BG1, BG2, BG3) containing 1, 2, and 3 objects respectively. Each background was used to test various versions of Stable Diffusion models: SD V1.2, SD V1.2 finetuned on inpainting tasks (ft-I), SD V1.4, and SD V1.4 finetuned on the COCO dataset with LoRa). The results demonstrate that SD V1.2 finetuned on inpainting tasks (ft-I) performs the best overall, producing better looking images that fit the requirments of the layout better. Notice the higher degree of physical consistency compared to the other models, which often generate objects that "phase out" or do not fit the background well.}
    \label{fig:veg_compare}
\end{figure}

\begin{table}[ht]
\centering
\begin{tabular}{|l|c|c|c|c|}
\hline
\multicolumn{5}{|c|}{\textbf{Images That Contained One or More Inpaint Failures}} \\ \hline
 & \textbf{SD V1.2} & \textbf{SD V1.2 ft-I} & \textbf{SD V1.4} & \textbf{SD 1.4 ft-cl} \\ \hline
BG1 (1 Object) & 3 & \textbf{0} & 5 & 2 \\ \hline
BG2 (2 Objects) & \textbf{1} & \textbf{1} & 4 & 4 \\ \hline
BG3 (3 Objects) & \textbf{2} & 7 & 6 & 7 \\ \hline
\textbf{Total} & \textbf{6} & 8 & 15 & 13 \\ \hline
\end{tabular}
\vspace{0.5cm}
\begin{tabular}{|l|c|c|c|c|}
\hline
\multicolumn{5}{|c|}{\textbf{Images That Contained One or More Clear Continuity Failures}} \\ \hline
 & \textbf{SD V1.2} & \textbf{SD V1.2 ft-I} & \textbf{SD V1.4} & \textbf{SD 1.4 ft-cl} \\ \hline
BG1 (1 Object) & 4 & \textbf{0} & 3 & 4 \\ \hline
BG2 (2 Objects) & 8 & \textbf{0} & 7 & 8 \\ \hline
BG3 (3 Objects) & 8 & \textbf{0} & 10 & 10 \\ \hline
\textbf{Total} & 20 & \textbf{0} & 20 & 22 \\ \hline
\end{tabular}
\caption{Results for the Stable Diffusion VEG comparison. (Top) How many of the examples (10 per background) for each SD model had at least one inpaint failure (with an object not being printed in any recognizable state). Version 1.2 and its finetuned variant fared the best. (Bottom) How many of the 10 examples for each background had at least one inpaint failure (with the physical continuity of the scene being broken in a clear and obvious way). The finetuned version 1.2 variant performed much better than any other. This version was chosen to represent Stable Diffusion in the VEG for all other experiments in this work.}
\label{tab:veg_compare_tab}
\end{table}

To determine the most effective version of the Stable Diffusion (SD) model for use as the Visual Element Generator (VEG) in further SPRING evaluation, we compared four versions: SD V1.2, SD V1.2 pre-finetuned on inpainting tasks (ft-I), SD V1.4, and SD V1.4 finetuned on the COCO dataset with LoRa (ft-cl). We tested each model across three backgrounds (BG1, BG2, BG3) with author-constructed layouts, containing 1, 2, and 3 objects respectively. Each completed image was checked for the presence of at least one inpaint failure or continuity failure. Inpaint failures in this case are defined as instances where an object is not generated in any recognizable state. Continuity failures are instances where the physical consistency of the scene is broken in a clear and obvious way. As this is a comparative test for choosing a base model, these metrics were applied rather generously, counting only breaches that were visually apparent as failures.

The results (Figure \ref{fig:veg_compare} and Table \ref{tab:veg_compare_tab}) showed that SD V1.2 finetuned on inpainting tasks (ft-I) consistently outperformed the other models. It exhibited the second fewest inpaint failures, with only 8 instances across all backgrounds, and no continuity failures at all. All objects generated during this experiment blended into the background quite convincingly. In contrast, the other models had significantly higher failure rates. Consequently, SD V1.2 ft-I was selected as the VEG backbone model for all subsequent experiments.
\end{rev}

\FloatBarrier
\subsection{Evaluating Image Quality}
\label{sec:exp_img_qual}

This experiment assesses the visual quality of images produced by SPRING using the leading automated image quality metrics. This measurement of visual quality is inherently tied to several implicit preferences in design, including aesthetics and naturalness.

\subsubsection{Setup}
\textbf{Generating Specifications.} To evaluate the visual quality of the produced images, 10,000 scenes were generated from SPRING and from baselines using random specifications generated via the following procedure. These specifications were unconstrained in position placement to focus on assessing the visual quality of images, not the feasibility of constraints in varied backgrounds.
Possible object types were selected based on the room being generated -- chair, couch, potted plant, dining table, and television for living rooms; microwave, oven, toaster, refrigerator, and potted plant for kitchens. A requested number of objects were selected for each scene, with replacement. Additionally, natural language prompts were generated programmatically for use with the Stable Diffusion baselines.
These prompts listed all objects to be included. For example, ``An oven, a refrigerator, and a toaster''.

\textbf{Baselines.} SPRING's performance was compared with Stable Diffusion, and SG2IM, representing text-to-image and graph-to-image methods respectively. Another text-to-image baseline -- GLIDE -- was also checked on a smaller set of examples, but was found to produce lower quality images than Stable Diffusion. Rectangular masks, centered in the image and featuring a 20-pixel margin around each edge, defined every mask in the Stable Diffusion baseline. This allowed Stable Diffusion to retain more context from the background.

During our evaluation, we found that the images generated by SG2IM were of low quality and displayed minimal variations when given the same scene graph. We maintained consistency with the parameters and code provided in the original SG2IM paper. 
After reaching out to the author via GitHub, we learned that the poor image quality was likely due to mode collapse. 
Nevertheless, we included these results in our evaluation for a comprehensive comparison.

\textbf{Background images.} Background images were generated by the algorithms themselves, increasing the difficulty of the task. SPRING handles this by using the VEG with the prompt ``a clean, empty, living room.'' or ``a clean, empty, kitchen.''. Stable Diffusion used the same approach, generating a background before receiving the specification and then inpainting it to complete the task. 
As SG2IM inherently generates its own backgrounds, this aspect remains consistent with its usual operation.

\textbf{Training.} SPRING was trained on a subset of the COCO 2017 Detection dataset relating to interior design. It is difficult to find real-world datasets of background images (containing few objects) with bounding boxes of the objects to be generated. We instead use datasets of images containing objects  and their corresponding bounding boxes (such as COCO). We remove these objects from the original images using the Telea method \cite{telea2004image}. This lets us train SPRING to place bounding boxes in their most natural positions. 
To ensure that the dataset was focused on interior design, only images including one or more of the following categories were included: chair, couch, potted plant, bed, mirror, dining table, window, desk, toilet, door, TV, microwave, oven, toaster, sink, refrigerator, or blender. In order to train SPRING in a supervised manner, it was necessary to remove these objects from the images, leaving only the empty backgrounds and bounding box annotations. To accomplish this, various methods were tested with the goal of achieving visually pleasing results while also being efficient as numerous images needed to be processed. The Telea method was ultimately selected as it uses an image's gradient information to guide the inpainting process, resulting in visually plausible results. 

SPRING was trained with a learning rate of 0.0001 and a batch size of 8 for 100 epochs. Negative Log Likelihood loss was utilized. Instead of training on pixel values, SPRING was trained on a relative system of per mille (parts per thousand). This allows the network to operate using the same parameters on multiple image sizes, but restricts precision slightly.

\textbf{Metrics.} The primary metrics we collect here are the Inception score (IS) \cite{salimans2016improved} and Fréchet Inception Distance (FID) \cite{NIPS2017_8a1d6947}. 
These are measures of the quality and diversity of images from a generative model based on the intermediate layers of a pre-trained classifier. For IS, higher scores indicate better quality images, while FID scores indicate better quality with lower scores. FID also requires comparison statistics to a control dataset -- in our case, COCO 2017. 
Both metrics are known to be volatile \cite{barratt2018note}, hence the large number of images produced for testing. 
Once compiled, all images are resized to $1000\times1000$ to make a fair comparison (IS and FID scores are effected by image size).

\begin{figure}[tp]
    \centering
    \includegraphics[width=0.78\linewidth]{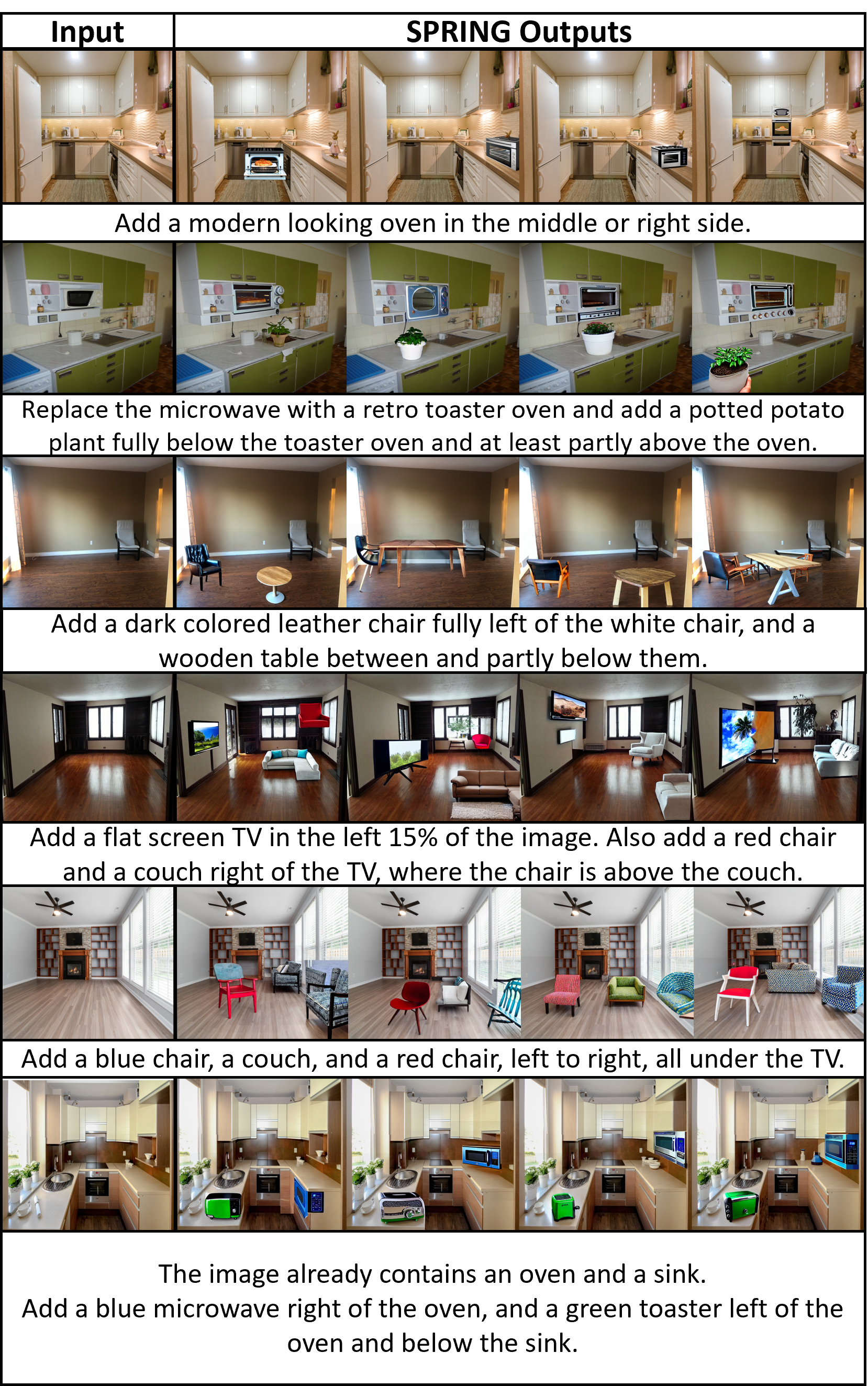}
    \caption{Example visualizations from SPRING. Note the diverse locations possible given each background and specification. Backgrounds are from various web-sources. Constraints are described in text for clarity. They were provided to SPRING in propositional logic (See appendix table \ref{tab:nl_to_log_1} for logic definitions).}
    \label{fig:pal1}
\end{figure}

\begin{figure}[tp]
    \centering
    \includegraphics[width=0.9\linewidth]{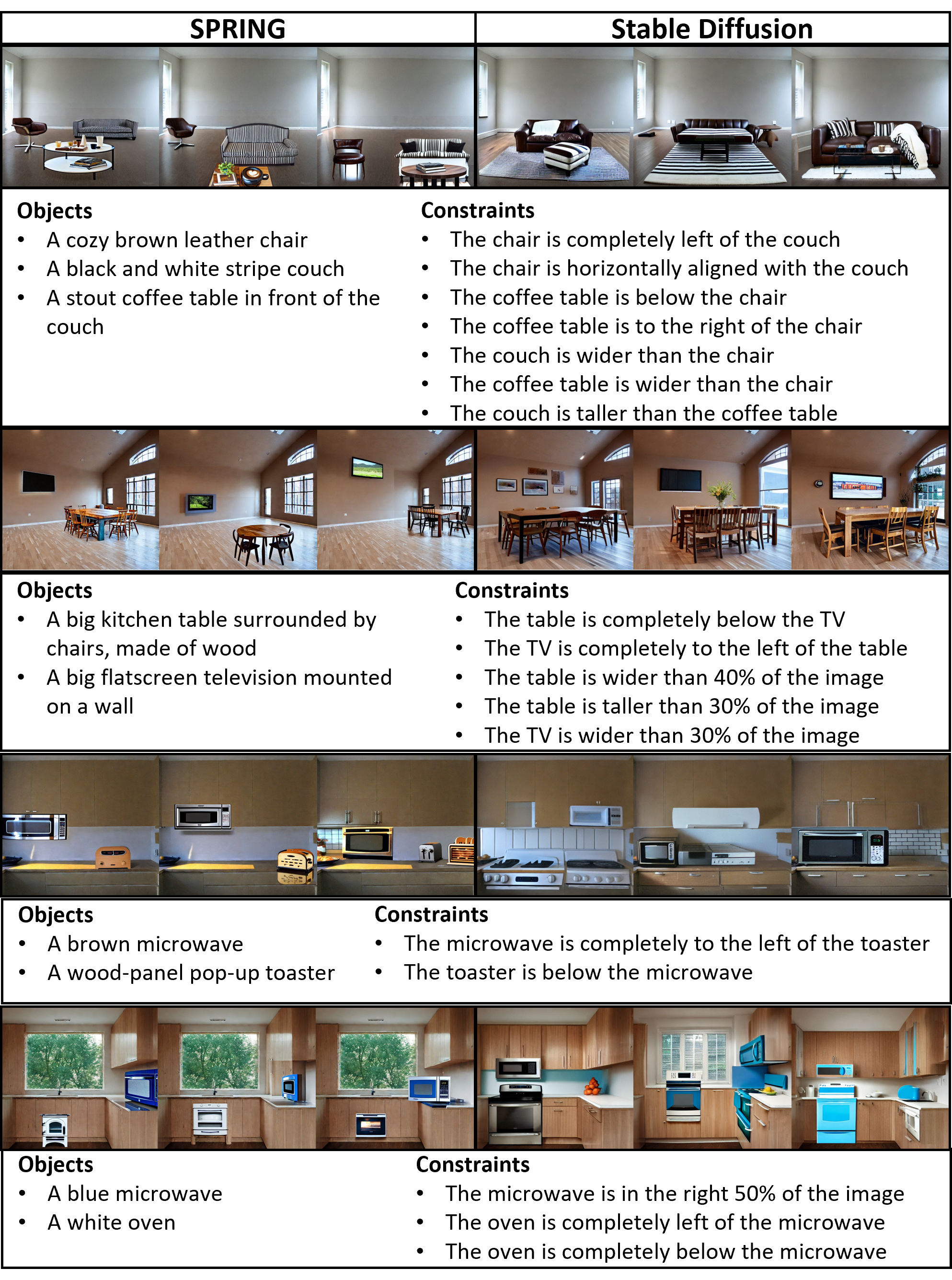}
    \caption{More visualizations from SPRING with comparisons to Stable Diffusion. Backgrounds are generated as detailed in section \ref{sec:exp_img_qual}. Constraints are described in text for clarity. They were provided to SPRING in propositional logic and Stable Diffusion as text (See appendix table \ref{tab:nl_to_log_2} for logic definitions).}
    \label{fig:pal2}
\end{figure}

\begin{figure*}[tb]
    \centering
    \small
    \includegraphics[width=0.95\linewidth]{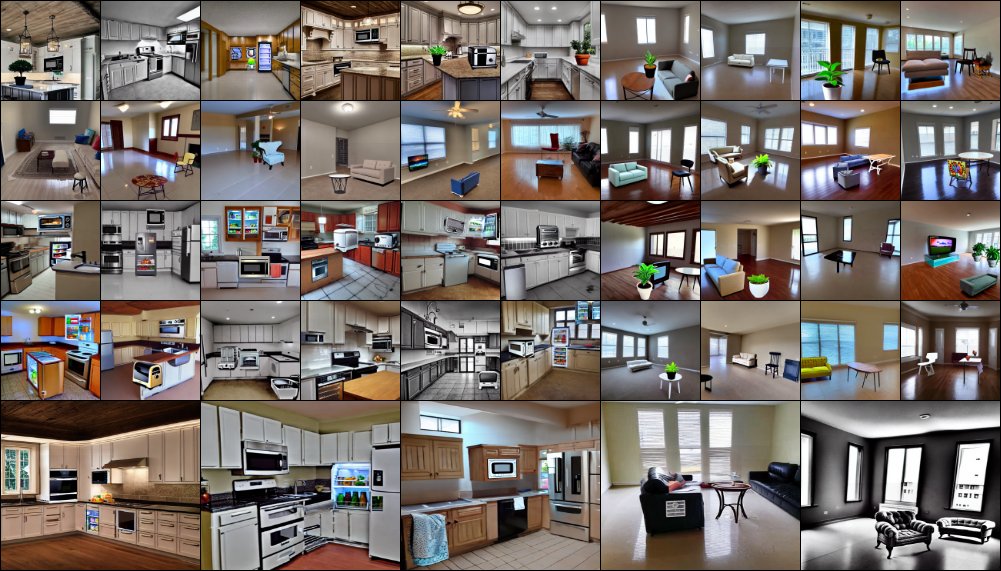}
    \vspace{-8pt}
    \caption{Images produced by SPRING with generated backgrounds and no positional constraints -- only objects are specified. SPRING's SRM still finds sensible locations for objects, even in the absence of complex constraints. Backgrounds are generated as detailed in section \ref{sec:exp_img_qual}. The list of objects generated for each image here can be found in \ref{sec:addi_exp_info_logiclang}. }
    \vspace{-12pt}
    \label{fig:extrapal}
\end{figure*}

\textbf{Figures.} From the 10,000 scenes created, a handful of them were chosen to represent this group in figure \ref{fig:extrapal}. In addition to the large set of randomized, unconstrained, generated-background designs, a small set of exemplar images were produced with author-constructed specifications both using real images from across the web (see figure \ref{fig:pal1}, \ref{fig:comppal}) and from generated backgrounds (see figure \ref{fig:pal2}). These could not feasibly be created at a scale where automated metrics like IS and FID would yield meaningful results, but they do allow an additional human-eye demonstration of image quality.

\subsubsection{Results}

\begin{table}[tb]
\centering
\small
{\fontsize{10pt}{12pt}\selectfont
\begin{tabular}{l|cc|ccc}
\hline
\textbf{Method} &
\textbf{Position Accuracy } $\uparrow$ &
\textbf{Object Accuracy } $\uparrow$ &
\textbf{IS} $\uparrow$ &
\textbf{FID} $\downarrow$ \\
\hline
Stable Diffusion   & 0.5          & 0.63          & 3.58 & 162.73 \\
SG2IM              & 0.0          & 0.0           & 2.52 & NaN \\
SPRING $\star$ & \textbf{1.0} & \textbf{0.77} & \textbf{3.59} & \textbf{160.36} \\
\hline
\end{tabular}
}
\caption{Quantitative results from 10,000 generated scenes show our SPRING (with Stable Diffusion in the VEG) leads in position and object accuracy for user specification satisfaction, as well as Inception Score (IS) and Fréchet Inception Distance (FID)  perception scores for visual image quality. 
FID statistics were calculated from the COCO 2017 validation set. Position accuracy represents how often the positional constraints are met. Object accuracy represents how often the correct number and type of objects are generated and identifiable. $\star$ indicates our method.}
\label{tab:img_qual}
\end{table}

Figure \ref{fig:extrapal} consists of selected scenes from the 10,000 automatically generated designs, while Figures \ref{fig:pal1} and \ref{fig:pal2} display some of the designs crafted from author-constructed specifications, with the former utilizing real-world images and the latter featuring VEG-generated backgrounds.
Quantitative results are presented in Table \ref{tab:img_qual}.

From these images, we can conclude that SPRING produces lifelike images.
We especially would like to draw our readers' attention to the images in Figure \ref{fig:rollout}. 
We see that the microwave and the toaster were added in reasonable locations that one would expect. 
Also, their dimensions and perspective are quite natural. 
We conclude that this is because the spatial reasoning module was able to discern the usual locations of these appliances, and it works in harmony with the visual element generation module, which was able to understand the context (such as the tabletop and the closet in the specific direction) and put the appliances in the most natural orientations. 
Neural generative models such as Stable Diffusion 
also generate realistic scenes, but their results often violate user specifications (see later sections). 
The image quality from SG2IM was the worst due to mode collapsing. 
Note that the images generated by SPRING do have occasional imperfections -- this suggests our future line of work. Most objects are placed realistically, but some are oriented oddly. We hypothesize that this issue is primarily due to the difficulty of learning locational preferences strongly from a limited dataset. The spatial reasoning module faces a complex task, working directly with an image feature vector to find suitable locations for every object, a challenge amplified in cramped scenes with strict constraints. Another contributing factor is the visual element generator, which has its own inductive biases. It often requires careful coaxing to generate furniture facing certain orientations, a likely consequence of its large-scale training on Internet data, where few images depict furniture from the back or facing away from the camera. Overall, though, the objects generated by SPRING are generally well-placed, integrating seamlessly into the given backgrounds. This implies that the SRM, despite the inherent difficulty of its task, performs well in its role while keeping constraints satisfied.

From the quantitative results, we can see that our assessment of image quality is validated at scale. Analyzing 10,000 generated scenes, SPRING’s performance is comparable to Stable Diffusion and higher than SG2IM in both FID and IS metrics. This indicates that the integration of our constraint satisfaction capability does not undermine the overall aesthetic quality of the images, as corroborated by these state-of-the-art metrics.

\FloatBarrier
\subsection{Evaluating Design Adherence}
\label{sec:exp_design_adh}
This experiment evaluates how well SPRING adheres to the user specification provided. It tests for explicit constraints around the spatial variables being upheld, as well as the correct objects being placed into the image.

\subsubsection{Setup}
\textbf{Generating Specifications.} 20 interior designs were produced for testing. Specifications were generated using the same process as the ``Evaluating Image Quality'' experiment (section \ref{sec:exp_img_qual}), except with constraints also being randomly generated. 
For ease of generation, only the constraints cleft, cright, cabove, and cbelow could be sampled -- indicating a non-overlapping directional constraint in the specified direction.
Each specification included 3 objects and 4 constraints, and was checked to ensure it was satisfiable. 
These constraints were also logged as scene graphs for use by SG2IM. The conversion from our language to scene graphs was done without ambiguity or information loss.

\textbf{Baselines.} The same baselines were tested as were introduced in the  ``Evaluating Image Quality'' experiment (section \ref{sec:exp_img_qual}) -- including Stable Diffusion and SG2IM. Natural language constraints were used to inpaint images using Stable Diffusion, these were formed from programmatically constructed prompts derived from the same logic as SPRING.  For SG2IM, which utilizes scene graphs as input, we converted these from SPRING's propositional logic without ambiguity.

\textbf{Background images.} Background images were generated using the same process as the 10,000-size set in the ``Evaluating Image Quality'' experiment (section \ref{sec:exp_img_qual}).

\textbf{Training. } Training was done using the same process as the ``Evaluating Image Quality'' experiment (section \ref{sec:exp_img_qual}).

\textbf{Metrics.} Object accuracy and position accuracy were measured as follows: \textit{Object accuracy.} the author counted and classified the objects in each image. If all 3 objects were visible and identifiable, the image was given a 100\% object accuracy. Two identifiable objects resulted in a 67\% accuracy, and so on. This score was then averaged across all 20 images to measure the object accuracy of the method. \textit{Position accuracy.} for each constraint clause (e.g. the oven is left of the toaster), the viewer checked whether the constraint was clearly true within the image. Clauses that referenced images which were not inpainted correctly (penalized in object accuracy) were not included in this count, only relationships between visible and recognizable objects. Position accuracy was also averaged across all 20 images to produce our final metrics.

\begin{figure*}[tp]
    \centering
    \includegraphics[width=\linewidth]{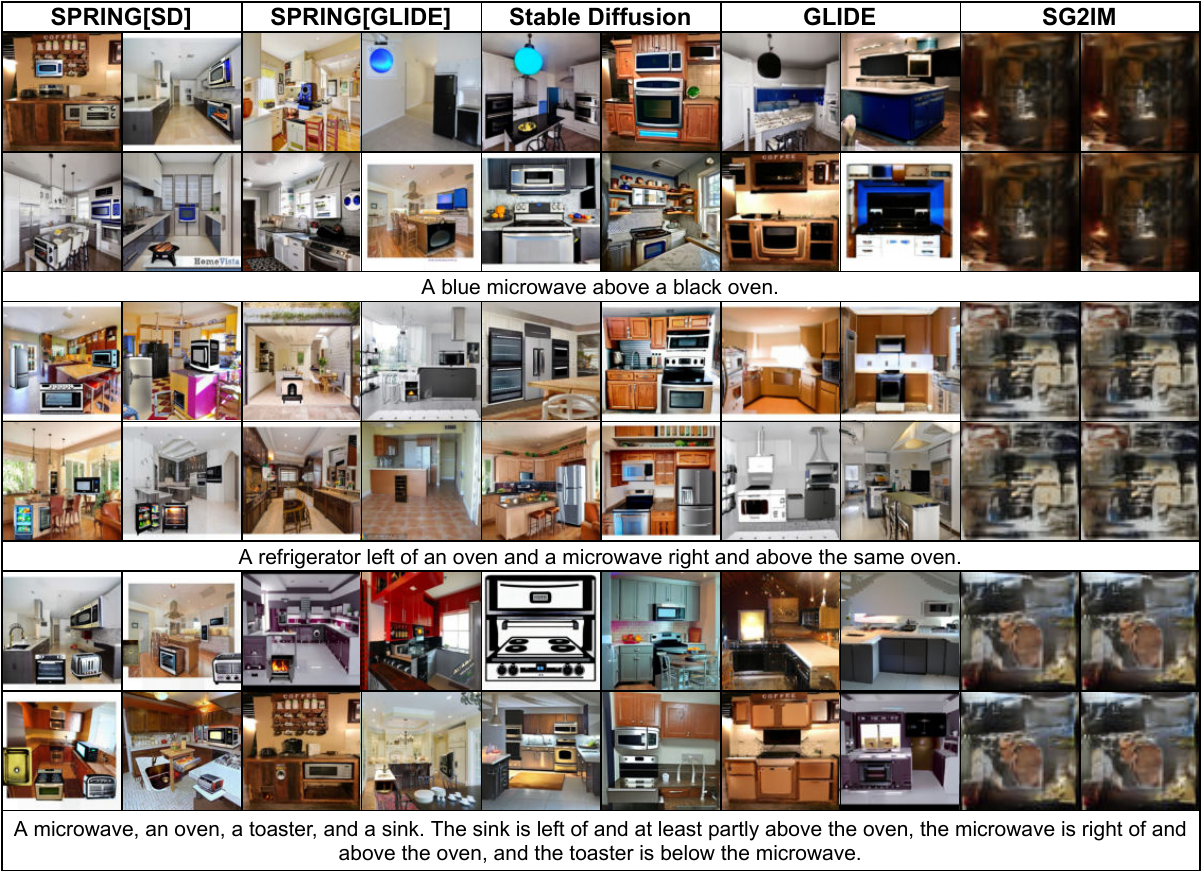}
    \vspace{-16pt}
    \caption{ Comparisons between SPRING (VEG choice in brackets) and baselines on kitchen backgrounds from various web-sources. SPRING creates distinct objects and always follows positional constraints. Baseline methods produce compelling images, but rarely fulfill all constraints. They also greatly disrupt the background. Notice how the word ``blue'' in the first row affects the SPRING and baseline images. SPRING applies blueness to the microwave, while Stable Diffusion and GLIDE apply it to other parts of the image. (See appendix table \ref{tab:nl_to_log_3} for logic definitions)}
    \vspace{-6pt}
    \label{fig:comppal}
\end{figure*}

\subsubsection{Results}
All metrics are presented in Table \ref{tab:img_qual}. Stable Diffusion had lower object accuracy than SPRING, and also showed much worse position accuracy compared to SPRING (50\% for Stable Diffusion alone compared to 100\% with SPRING). Identifying  objects was not possible with SG2IM, hence its accuracy is 0. Notice that SPRING does not guarantee perfect object accuracy, but improves it by explicitly painting one object at a time in isolation from others. SPRING does guarantee perfect position accuracy via the symbolic constraint reasoning of the sampling procedure in the SRM. 
Several failing images of Stable Diffusion suggest the struggle of current neural generative models. 
For example, in the top row of Figure \ref{fig:comppal}, Stable Diffusion was able to understand that some blue objects need to be added. However, it cannot carry out deep semantic reasoning to determine that the blue color is associated with the microwave. This same failing is visible in the right panel of Figure \ref{fig:functionality}, where the descriptive texts ``blue'' and ``green'' are applied widely and arbitrarily, instead of being applied to their respective objects.

\FloatBarrier
\subsection{Evaluating Spatial Reasoning}
\label{sec:exp_srm}
\subsubsection{Setup}
This experiment assesses how well the generated object locations satisfy explicit constraints and implicit preferences in a controlled setting.
This is done by creating synthetic datasets which have both explicit constraints and implicit preferences. 
To measure implicit preferences, the objects' positions are biased in ways unknown to the algorithms tested, but known to the authors.
The goal is then to examine which algorithm can best capture this implicit bias (i.e., putting objects in locations following the inductive biases present in the training dataset). 
Explicit constraints are measured by checking if the sampled locations are properly constrained after execution.
To eliminate outside variables, this section focuses on the spatial reasoning module in isolation, testing its ability to learn ``good'' positions while guaranteeing that all spatial constraints are satisfied. 
\begin{rev}[17]
In a broader scope, these implicit preferences may include aesthetic considerations related to natural and visually appealing placement of objects within the scene.
\end{rev}

\textbf{Dataset \& Metrics.} We define three synthetic scenarios with known object location distributions, known to the synthetic data generator but not the algorithms. No VEG or perception module is used. We evaluate two key metrics: \textit{constraint accuracy} and \textit{preference accuracy}. \textit{Constraint accuracy} measures the percentage of generated locations that adhere to the explicitly provided constraints essential for the given scenario. On the other hand, \textit{preference accuracy} quantifies the extent to which generated locations align with the learned implicit preferences. See \ref{sec:syn_ds} for scenario specifications, including the list of explicit constraints required and the implicit preferences to be checked. 

\textbf{Baseline.} For qualitative comparisons on the spatial reasoning module, we compare with baselines that also produce bounding boxes for objects to be generated. 
Unfortunately, conditional neural generative models, such as Stable Diffusion, cannot be compared fairly because they do not provide coordinates for the generated objects. 
Hence, we compare SPRING planning with a Generative Adversarial Network (GAN)  augmented with a differentiable convex optimization (CVX) layer. 
More specifically, this approach models the locations of the generated objects as optimal solutions to Quadratic Programs (QPs) subject to constraints. 
The objective function of the QP is designed to capture implicit location preferences and is learned via backpropagation (the work of \cite{cvxpylayers} details an algorithm to backpropagate gradients over a convex optimization layer). 
To be specific, the objective function of the QP is updated during training to match the QP's optimal solution (predicted locations) with the actual locations in the dataset. 
Notice that the ``GAN + CVX'' approach can only deal with convex constraints and objective functions, which limits its applicability.

\textbf{Training.} Both methods are trained on each scenario until convergence, and each dataset is matched with a set of constraints during testing.

\subsubsection{Results}

\begin{table}[tb]
\centering
\small
{\fontsize{10pt}{12pt}\selectfont
\begin{tabular}{c|cc|cc}
\hline
\textbf{Scenario} &
\multicolumn{2}{c}{\textbf{SPRING $\star$}} &
\multicolumn{2}{c}{\textbf{GAN+CVX}} \\
&
\multicolumn{1}{c}{\textbf{Pref. Acc. $\uparrow$}} &
\multicolumn{1}{c}{\textbf{Constr. Acc. $\uparrow$}} &
\multicolumn{1}{c}{\textbf{Pref. Acc. $\uparrow$}} &
\multicolumn{1}{c}{\textbf{Constr. Acc. $\uparrow$}} \\ \hline
``Basic'' & \textbf{1.0} & 1.0 & 1.0 & 1.0 \\ \hline 
``Tight'' & \textbf{0.917} & 1.0 & 0.813 & 1.0 \\ \hline
``Complex'' & \textbf{0.875} & 1.0 & 0.723 & 1.0 \\ \hline
\end{tabular}
}
\caption{Results for evaluating ``goodness'' of location decisions by the SRM when tested on synthetic scenarios. Our SPRING is able to generate object locations better satisfying implicit preferences from data (measured by preference accuracy) compared to baselines, while both approaches 100\% satisfy explicit user constraints. Note, SPRING can also handle non-convex constraints, while ``GAN+CVX'' cannot. More details of this experiment are in \ref{sec:syn_ds}. $\star$ indicates our method.
}
\label{tab:loc_qual}
\end{table}

As shown in Table \ref{tab:loc_qual}, the SRM formulation SPRING uses outperformed or matched GAN + CVX in preference accuracy for every dataset. Both algorithms guarantee positional constraints, but GAN + CVX only does this with convex constraints. This implies that the SRM used by SPRING is very capable of learning implicit preferences from data while still maintaining complex spatial constraints. Both approaches were designed to satisfy the explicit constraints with 100\% accuracy, and they did.

\FloatBarrier
\subsection{Demonstrating Zero-shot Constraint Satisfaction}

\begin{figure*}[tb]
    \centering
    \small
    \includegraphics[width=0.9\linewidth]{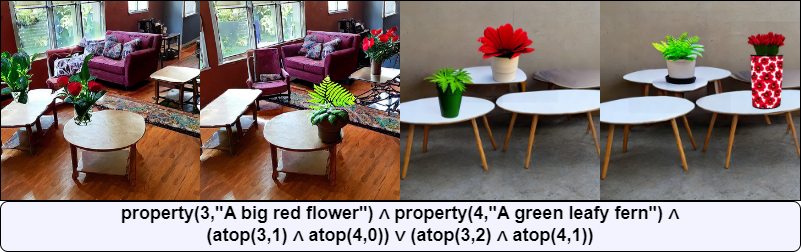}
    \vspace{-8pt}
    \caption{SPRING's zero-shot transfer learning performance on a new constraint \texttt{atop} never present in the training set. 
    \texttt{atop} constrains an object to be sitting on top of another. This specification applies to all backgrounds which already have three tables: 0, 1, and 2 from left to right. The flower is always one table to the right of the green fern, showing good performance of SPRING.}
    \vspace{-12pt}
    \label{fig:zeroshot}
\end{figure*}

Our approach allows for re-programming of the spatial reasoning module for zero-shot constraint editing. New constraints can be implemented and used in filtering the GRU outputs of the SRM at any time, even after training. To demonstrate this capability, we construct a new constraint type after training SPRING. We define this constraint as \texttt{atop($o_1, o_2$)} -- which constrains the first object's bounding box ($o_1$) to be on top of the first object's bounding box ($o_2$). More precisely, \texttt{atop} evaluates to true if and only if:
\begin{itemize}
    \item the bottom of the first object's bounding box is between the top and the bottom of the second object's bounding box.
    \item the right side of the first object's bounding box is between the right and left sides of the second object's bounding box.
    \item the top of the first object's bounding box is above the top of the second object's bounding box.
\end{itemize}
By combining this new constraint with the previous language set, very complex specifications can be produced. In Figure \ref{fig:zeroshot} in the main text, \texttt{atop} is used to define a complex specification utilizing the perception module to create a design which can be applied to any background including three tables. A flower and a fern are to be placed in the image. The flower is always one table to the right of the green fern (e.g. if the fern is on the middle table, the flower is on the right table, and if the fern is on the left table, the flower is on the middle table). As demonstrated in that figure, specifications can be automatically applied to entire sets of backgrounds, and produce diverse and complex images through use of zero-shot constrains.

\FloatBarrier
\begin{rev}[16]
\subsection{Demonstrating Extensibility of SPRING Outside of Interior Design}
\label{sec:extens}

\begin{figure*}[tb]
    \centering
    \includegraphics[width=\linewidth]{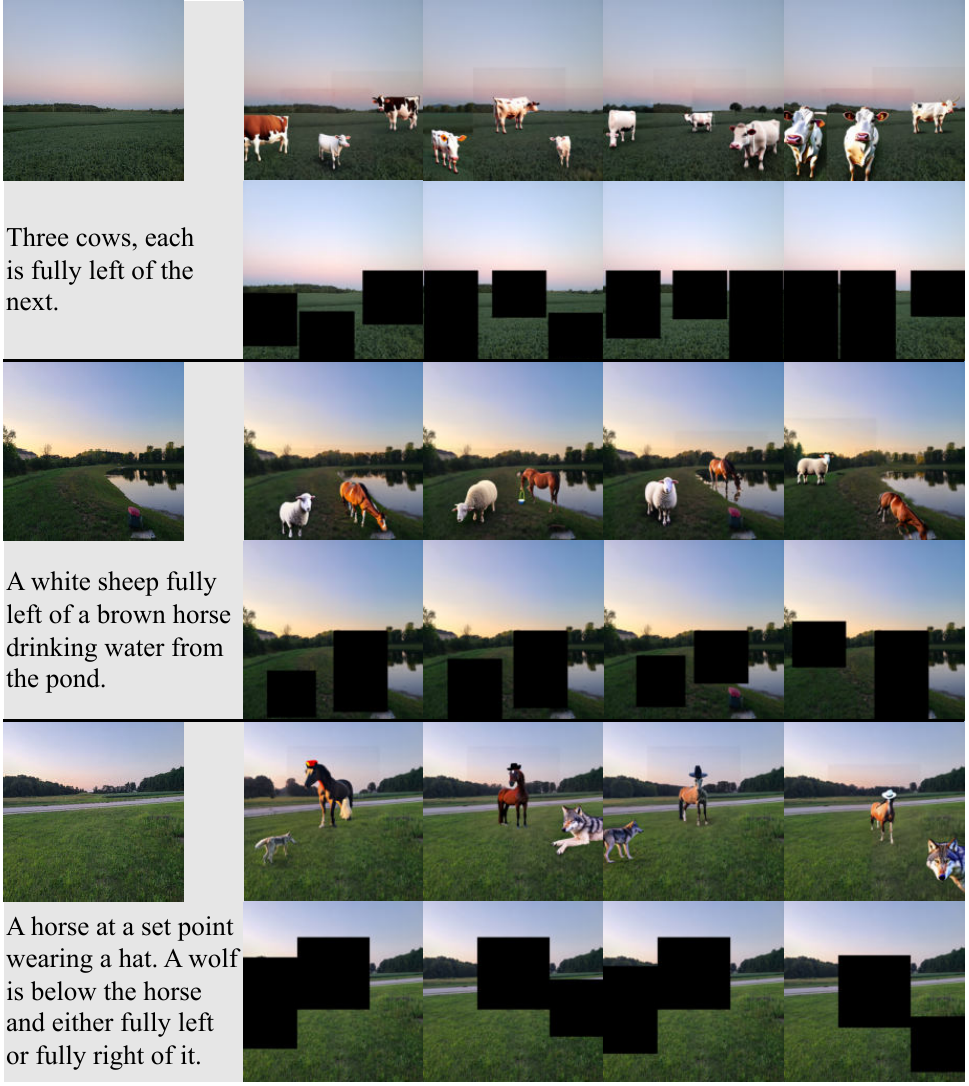}
    \vspace{-10pt}
    \caption{SPRING results on an ``animals'' subsection of the COCO dataset. This included all training set images that included one or more of the following categories: dog, horse, sheep, cow, elephant, bear, giraffe. These results show that SPRING can be easily extended to other image-generation domains. \vspace{-8pt}}
    \label{fig:animals}
\end{figure*}

In the prior experiments, we demonstrated the effectiveness of SPRING in generating aesthetically pleasing and highly constrained interior design images. We chose the domain of interior design as it perfectly aligns with SPRING’s strengths—requiring both implicit knowledge gleaned from data for aesthetic and common-sense adherence, and explicit knowledge from users specifying design requirements that must be meticulously followed. However, the potential applications of SPRING extend far beyond interior design. Figure \ref{fig:animals} illustrates an example of SPRING applied to a different subset of the COCO dataset, this time featuring animals instead of furniture elements. The results maintain high quality and adherence to specified input constraints, showcasing SPRING’s versatility.

SPRING can be applied across any domain that demands stringent 2D constrained image generation. The only prerequisite is a robust dataset of labeled images. The Perception module can either be trained afresh on an object detection network, as was done with DETR, or leverage a pre-trained model. Meanwhile, the Visual Element Generator requires an inpainting network, which, although challenging and resource-intensive to train from scratch, can be efficiently fine-tuned or used as-is, as demonstrated in this study. The Spatial Reasoning Module necessitates only a moderate dataset of images labeled with object categories and bounding boxes to learn implicit locational preferences. 

With the zero-shot extensibility of the constraint language included, SPRING can be adapted to new domains easily. Interior design was selected as an illustrative domain not only because it is a natural fit for SPRING's capabilities but also due to the ready availability of necessary data and its practical relevance. 

\end{rev}

\FloatBarrier
\subsection{Human Study}
\label{sec:human_study}
\begin{table}[tbh]
\centering
\small
{\fontsize{10pt}{12pt}\selectfont
\begin{tabular}{l|cccc}
\hline
\multicolumn{1}{c|}{\textbf{Method}} &
  \multicolumn{1}{c}{\textbf{\begin{tabular}[c]{@{}c@{}}Specification\\ Satisfaction $\uparrow$\end{tabular}}} &
  \multicolumn{1}{c}{\textbf{\begin{tabular}[c]{@{}c@{}}Aesthetics $\uparrow$\end{tabular}}} &
  \multicolumn{1}{c}{\textbf{\begin{tabular}[c]{@{}c@{}}Background\\ Preservation $\uparrow$\end{tabular}}} &
  \multicolumn{1}{c}{\textbf{\begin{tabular}[c]{@{}c@{}}Spatial\\ Naturalness $\uparrow$\end{tabular}}} \\ \hline
SPRING $\star$           &   \textbf{4.068}    &    3.367   &   \textbf{4.473}    &   3.663    \\
Stable Diff. &    2.240   &    \textbf{3.588}   &   2.608    &    \textbf{3.876}   \\ \hline
\end{tabular}
}
\caption{Results for the human survey with two methods: Stable Diffusion and SPRING (ours). Metrics collected include Likert scores (1-5) for specification satisfaction, aesthetic appeal, background preservation, and the naturalness of spatial qualities -- e.g. proportions between objects, proportions between an object and the background. SPRING does much better at satisfying user specifications than Stable Diffusion, and is far less likely to damage the background due to its more controlled inpainting. SPRING receives slightly lower results for aesthetics and naturalness, perhaps representing a small trade-off between the amount of control provided by the algorithm and its ability to produce good-looking images. $\star$ indicates our method.}
\label{tab:human_exp}
\end{table}

This section presents a human study conducted to evaluate the performance of our SPRING method and Stable Diffusion inpainting for generating interior space images, such as kitchens, living rooms, and billiard rooms. We decided to conduct this study due to the difficulty of evaluating generated content for qualities like aesthetic appeal. Despite the existence of metrics like IS and FID used during the automated experiments, human studies remain the gold standard in generated image evaluation.

Participants were provided with a background image, a completed scene, and a bullet-pointed specification used to generate the image. Additionally, questions with images generated by SPRING displayed the intermediate layout. Background images were generated using the methods described in \ref{sec:exp_img_qual}. Constraints were handcrafted by the authors to fit the background. This was done with the aim of evaluating all approaches in good conditions -- so that all algorithms can produce the highest quality images possible.

\subsubsection{Evaluation Metrics}
\label{sec:study_evaluation_metrics}

\begin{rev}[9]
Participants were asked to rate each image on a 5-point Likert scale across four distinct metrics: specification satisfaction, aesthetic appeal, background preservation, and spatial naturalness. Detailed metric descriptions can be found in \ref{sec:addi_exp_info_questiontext}.

\begin{itemize}
\item \textbf{Specification satisfaction:} Image adherence to the input specifications, including object visibility, constraint satisfaction, and details like color, texture, and material.
\item \textbf{Aesthetic appeal:} Overall image quality, irrespective of specification.
\item \textbf{Background preservation:} Maintenance of the original background without unnecessary changes.
\item \textbf{Spatial naturalness:} Reasonableness of object locations and dimensions within the image.
\end{itemize}
\end{rev}

\subsubsection{Questionnaire Structure}
\label{sec:questionnaire_structure}

The questionnaire included 3–4 sections, with versions 1 and 2 having three sections and version 3 having four sections. Versions were randomly assigned to each participant. Each section contained six examples generated by both SPRING and Stable Diffusion, using the same constraints and backgrounds. To ensure the validity of the responses, two ``attention check / gold-standard'' questions were incorporated into each version of the questionnaire. These questions have obvious answers for human beings. Responses that violated these attention checks were discarded. A total of 9 human subjects participated in the survey.

The specification satisfaction part is a multifaceted question involving some objectivity (are the positional constraints satisfied, are the right number of images displayed) and some subjectivity (do the objects look like the types they are supposed to be, do they have the right texture or material). For this reason, the specification satisfaction Likert score question is preceded by two additional questions where the subject is asked to check the boxes for the objects that are not visible and the constraints that are not satisfied. This means the subject is forced to recognize the objective parts of specification satisfaction before taking into account the subjective ones.

\subsubsection{Initial Tutorial}
\label{sec:initial_tutorial}

Before the test could be completed, an initial tutorial was provided to the participants. A real image was presented, and the subjects were asked to fill out the questions in the same way as they would in the main study. After completing the example, the subject was given example answers so that they could check their understanding of the directions before they began the test.

\subsubsection{Demographics}
\label{sec:demographics}

\begin{rev}[10]
The study included participants with diverse backgrounds to ensure a comprehensive evaluation. Details can be found in \ref{sec:addi_exp_info_demographics}.
\end{rev}

\subsubsection{Results}
The averaged results are presented in Table \ref{tab:human_exp}. Our findings indicate that SPRING nearly matches the aesthetic quality and spatial naturalness of Stable Diffusion, as anticipated. Furthermore, SPRING significantly outperforms Stable Diffusion in terms of specification satisfaction and background preservation. These results demonstrate the effectiveness of the SPRING method for generating high-quality interior space images that adhere closely to the provided specifications while maintaining the integrity of the original background. These results support those found in the previous experiments: \textbf{better specification satisfaction with comparable or superior image quality.}

\begin{rev}[14]
\subsection{Computational Resources}
\label{sec:resources}
In training for spatial reasoning, the SPRING instances utilized here comprise 15,721,089 trainable parameters, spanning both the SRM and the trainable segments of the Perception module, excluding the VEG. Utilizing a modest GPU (NVIDIA GeForce RTX 2070 Super), training averaged 1.8 seconds per batch of eight images.

Evaluating efficiency outside the training phase is more nuanced due to the variable iterations of the GRU, which depend on the number of objects, constraint behaviors, and implicitly learned preferences. It is challenging to precisely describe SPRING's runtime in general scenarios. However, it is known that single forward passes of the GRU take approximately 0.5 milliseconds. During iterative refinement—with three potential actions and a fixed maximum decision depth determined by the integer per-mille range of each variable, the worst-case number of calls to the RNN can grow exponentially with the number of objects -- the typical time characterization for a depth first search.

The runtime of the VEG varies based on several factors, notably the choice of backbone model (e.g., Stable Diffusion, GLIDE). These models' efficiency has been thoroughly analyzed in prior studies. Importantly, in all evaluations conducted for this study, locating all objects with the SRM consistently required significantly less time than it took for a Stable Diffusion-based VEG to inpaint a single object into the scene, which averaged 12 seconds per inpaint. In other words, augmenting any inpainting model with the capabilities of SPRING is not a significant tradeoff for speed.

\end{rev}

\FloatBarrier

\begin{figure}[t]
    \centering
    \includegraphics[width=0.94\linewidth]{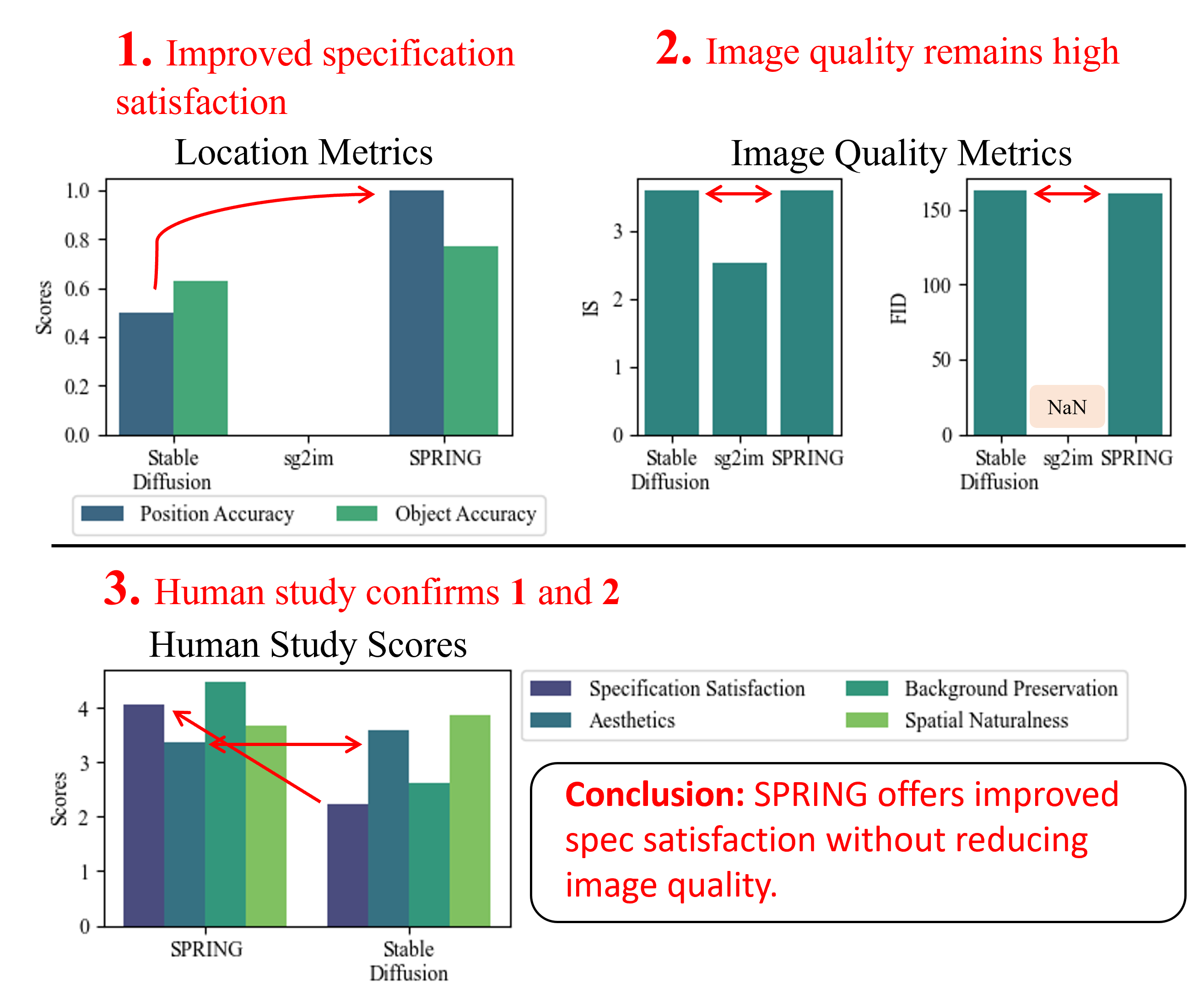}
    \caption{A bar-chart summary of evaluation metrics. Note the 3 main points: SPRING is better at specification satisfaction, SPRING creates images approximately as good-looking as state-of-the-art image generators (Stable Diffusion), and the human study supports both of the previous points. Background preservation and spatial naturalness scores are also given as specific aspects of specification satisfaction and aesthetics respectively. The pattern holds for these as well -- similar naturalness, improved preservation. }
    \label{fig:barcharts}
\end{figure}

\begin{rev}[13]
\subsection{Key Experimental Conclusions}
\label{sec:key_conc}

Drawing from the results of all of these experiments, several key conclusions can be drawn:
\begin{itemize}
    \item \textbf{SPRING is better than baselines at producing design scenes that fit the specification}: This is evident from the design adherence experiments (higher position accuracy and object accuracy), as well as the scores in the human study. See Figure \ref{fig:barcharts}.
    \item \textbf{SPRING has comparable or superior image quality to baselines}: This is evident from the image quality experiments (similar or better FID and IS scores), the human study, and the targeted spatial reasoning experiment (superior preference accuracy). See Figure \ref{fig:barcharts}.
    \item \textbf{SPRING can create complex scenes with zero-shot adapted constraints}: This is demonstrated in the creation of the \texttt{atop} constraint and the images combining it with conjunctions and disjunctions. See Figure \ref{fig:zeroshot}.
    \item \textbf{SPRING can be extended to other domains}: SPRING was easily extended to the domain of outdoor animal images as an exemplar case. See Figure \ref{fig:animals}.
\end{itemize}

\end{rev}

\FloatBarrier
\section{Conclusion}

We have introduced SPRING for design production. Good designs must meet specific user requirements and comply with implicit guidelines on appearance and functionality. This is only feasible through the \textbf{union of symbolic reasoning with neural algorithms}. We achieve this by integrating data-driven neural generative models with symbolically-driven constraint programming. The object positions suggested by the neural network are filtered through the symbolic constraint reasoning process to meet user specifications. 
Our SPRING produces high-quality interior design scenes that respect user-defined spatial requirements, satisfy common sense and look pleasing. 
Our approach also handles new constraint types in a zero-shot manner due to the symbolic constraint reasoner's programmability. It is also more interpretable as decisions are made iteratively.

\begin{rev}[18]
One limitation of SPRING is the expressibility of preferences in relation to constraints. Constraints are defined explicitly and concretely. For example, $\texttt{above}(o_1, o_2)$ resolves true if and only if object 1 is above object 2, whether it's by a single pixel or a greater distance. While our constraint language does support offsets to handle this (e.g., $\texttt{above}(o_1, o_2, 100)$ ensures object 1 is above object 2 by at least 100 per-mille), constraints are not inherently adjusted based on implicit user preferences. This is not to say that implicit learned preferences or ``fuzzy'' knowledge play no role in object positioning. In fact, preferences related to object types and the background image influence the iterative refinement process, determining final placements. However, the direct modification of constraints based on user preferences remains an area for future research. Nevertheless, SPRING mitigates the effects of this limitation with an optional human feedback step, allowing users to adjust constraints based on initial results. 

Another limitation of our work is the dependence on the VEG model's quality and compatibility for object generation. While the backbone VEG model can be finetuned by various effective methods, this is an expensive and difficult process that still offers limited control. New methods are in development that may reduce the severity of this limitation considerably by introducing more controllable generators \cite{zhang2023adding, mo2023freecontrol} -- potentially suitable for integration with SPRING.

With this in mind, future work will focus on enhancing control between the SRM and VEG and adapting to other domains. It will also include more exploration into user interaction with the SRM, potentially introducing softer constraints that can adapt to user feedback efficiently. Additionally, this method should naturally be expanded to cover more complex cases, such as 3D design.
\end{rev}

\section*{Acknowledgements}
This research was supported by
NSF grants IIS-1850243, CCF-1918327.





\bibliographystyle{elsarticle-harv} 
\bibliography{main}

\newpage
\appendix
\onecolumn

\section{Constraint Checking and Pruning}
\label{sec:con_check_prune}

The following section details constraint checking and pruning within the SRM's sampling procedure.

\textbf{Constraint Representation.} Consider the constraint language input (See Section \ref{sec:design_lang}) as an abstract syntax tree (AST) $c$. To avoid confusion with the search tree, the terms ast-root, ast-node, ast-leaf, etc, will be substituted for root, node, and leaf (which will refer exclusively to the search tree). The ast-root of $c$ will usually be an AND term, acting as a list of top-level constraints. AND, OR, and NOT terms all act as internal ast-nodes, while spatial relations act as ast-leaves. 

\textbf{Checking Constraints at Leaf Nodes.} At a leaf node, checking constraint satisfaction is trivial. First, the ast-tree representing the constraints in the spec is parsed using recursive descent to handle Structural terms like AND ($\wedge$) and OR ($\vee$).  If the term is an AND, SPRING recurses on the term to get a new list of ast-child terms, all of which must still hold for the node to be satisfiable. Conversely, if the term is an OR, we must recurse with the flag that the entire term resolves true if any of its ast-children do. Each spatial constraint --  like $\texttt{left\_of}$, $\texttt{above}$, etc -- is converted to its parameter-driven form. A full breakdown of these transformations to meaningful parameter constraints is described in Table \ref{tab:spatial-relations}. These parameter constraints can be checked directly against a leaf node range table, as the table is definitionally deterministic. This simple checking at leaf nodes would be enough for SPRING to function, but may be inefficient on its own, as many branches of the tree are clearly unsatisfiable.

\textbf{Range-based Pruning.} To decrease the search space further and avoid exploring clearly unsatisfiable regions, SPRING incorporates range-based pruning. Given the prospective range table and an ast-root term in the constraint language, we check each term that is an ast-child of the ast-root. If the term is a spatial relationship (such as $\texttt{left\_of}$, $\texttt{above}$, $\texttt{wider}$, $\texttt{zeq}$, etc.), it can be checked for satisfaction directly against the range table. For example, if the term $\texttt{left\_of}(o_1, o_2, 100)$ is found, we must ensure that object 1's x-value is smaller than object 2's x-value by at least 100 per-mille. In other words, the minimum of the ($o_1$, x) range must be less than the maximum of the ($o_2$, x) range minus the offset of 100. If this is not true, the current range table is unsatisfiable and its node can be marked as violating. That node and all of its children can be pruned. This can also be seen in Figure \ref{fig:srm_search}, step 2, where a decision is pruned for violating a constraint.

\section{General Additional Experiment Details}
\label{sec:addi_exp_info}

\subsection{Logical Language Translation}
\label{sec:addi_exp_info_logiclang}
The logical instructions for generating the figures in this work are given in the following tables. Note the relation \texttt{default}, which combines a number of default constraints used on all objects to fit the limitations of the VEG module. For the figures generated using Stable Diffusion as the VEG, the objects are set to have a width and height between 256 and 512 pixels. Additionally, the bounding boxes for these objects are constrained such that they do not extend beyond the boundaries of the image. For the figures generated using GLIDE as VEG, the same default constraints are used, with the exception that the objects are bounded between 150 and 256 pixels. This does not include Figure \ref{fig:extrapal}, which has its objects described for each image in Table \ref{tab:extrapal_objs} -- as that figure does not include positional constraints beyond the default.

\begin{rev}[9]
\subsection{Human Study Question Text}
\label{sec:addi_exp_info_questiontext}
This section provides the original questions associated with each rating metric in the human study, offering insight into the specific aspects participants were instructed to consider while assessing the images. 

\begin{itemize}
\item Specification satisfaction: "On a scale of 1 to 5, how well does the image fit the prompt? This includes the objects being visible, the constraints being satisfied, and the details mentioned in the prompts (e.g., objects being the right color). Do not take image quality into account."
\item Aesthetic appeal: "On a scale of 1 to 5, rate the aesthetic quality of the image. This includes how realistic the image is, how good each object looks, and how good the scene as a whole looks. Do not take the prompt into account."
\item Background preservation: "On a scale of 1 to 5, rate how well the background image was preserved. This includes unnecessary changes being made to the background not specified by the prompt, additional objects being added, details being unnecessarily removed, etc."
\item Spatial naturalness: "On a scale of 1 to 5, how reasonable are the locations and dimensions of the objects in the image."
\end{itemize}
\end{rev}

\begin{rev}[10]
\subsection{Human Study Demographics}
\label{sec:addi_exp_info_demographics}
All subjects were college graduates with degrees in science or technology fields. Most participants had some knowledge and familiarity with generative models or AI-generated content. All subjects are currently living in the United States of America, but have diverse countries of origin. All subjects understand the English language to an advanced degree, but for some, it is a second or third language. No subjects are known to have experience in interior design. Roughly half of the subjects are currently pursuing careers in academia, while the other half are pursuing careers in industry. 
\end{rev}

\begin{table}[H]
\resizebox{\textwidth}{!}{%
\begin{tabular}{l|l|l}
\textbf{Figure} & \multicolumn{1}{c|}{\textbf{Logic Language}} &
  \multicolumn{1}{c}{\textbf{Natural Language}} \\ \hline
Fig. \ref{fig:pal1} & \begin{tabular}[c]{@{}l@{}}type(0,"oven") $\wedge$\\ property(0,"a modern looking oven") $\wedge$ \\ default(0) $\wedge$ \\ right\_value(0, 400)\end{tabular} &
Add a modern looking oven in the middle or right side. \\ \hline
Fig. \ref{fig:pal1} & \begin{tabular}[c]{@{}l@{}}\# 0: a microwave, 1: an oven.\\ type(2,"toaster") $\wedge$ type(3,"plant") $\wedge$\\ property(2,"a retro toaster oven") $\wedge$\\ property(3,"a potted potato plant") $\wedge$ \\ default(2) $\wedge$ default(3) $\wedge$ \\ xeq(2,0) $\wedge$  yeq(2,0) $\wedge$ \\ wider(2,0) $\wedge$  taller(2,0) $\wedge$ \\ cbelow(3,0)  above(3,1)\end{tabular} &
\begin{tabular}[c]{@{}l@{}}Replace the microwave with a retro toaster oven \\ and add a potted potato plant fully below the toaster oven \\ and at least partly above the oven.\end{tabular} \\ \hline
Fig. \ref{fig:pal1} & \begin{tabular}[c]{@{}l@{}}\# 0: a chair.\\ type(1,"chair") $\wedge$ type(2,"table") $\wedge$\\ property(1,"a dark colored leather chair") $\wedge$\\ property(2,"a wooden table") $\wedge$\\ default(1) $\wedge$ default(2) $\wedge$\\ cleft(1,0,300) $\wedge$ below(2,1) $\wedge$\\ below(2,0) $\wedge$ left(2,0) $\wedge$ right(2,1,100)\end{tabular} &
\begin{tabular}[c]{@{}l@{}}Add a dark colored leather chair \\ fully left of the white chair, \\ and a wooden table between and below them.\end{tabular} \\ \hline
Fig. \ref{fig:pal1} & \begin{tabular}[c]{@{}l@{}}type(0,"tv") $\wedge$ type(1,"couch") $\wedge$ type(2,"chair") $\wedge$\\ property(0,"a flat screen tv") $\wedge$\\ property(1,"a couch facing left") $\wedge$\\ property(2,"a chair") $\wedge$\\ default(0) $\wedge$ default(1) $\wedge$ default(2) $\wedge$\\ left\_value(0,150) $\wedge$\\ cright(1,0) $\wedge$ cright(2,0) $\wedge$ cabove(2,1)\end{tabular} &
\begin{tabular}[c]{@{}l@{}}Add a flat screen TV in the left 15\% of the image. \\ Also add a red chair and a couch right of the TV, \\ where the chair is above the couch.\end{tabular} \\ \hline
Fig. \ref{fig:pal1} & \begin{tabular}[c]{@{}l@{}}\# 0: a tv.\\ type(1,"chair") $\wedge$ type(2,"couch") $\wedge$ type(3,"chair") $\wedge$\\ property(1,"a red chair facing away") $\wedge$\\ property(2,"a couch facing away") $\wedge$\\ property(3,"a blue chair facing away") $\wedge$\\ default(1) $\wedge$ default(2) $\wedge$ default(3) $\wedge$\\ cbelow(1,0) $\wedge$ yeq(2,1) $\wedge$ yeq(3,1) $\wedge$\\ cleft(1,2) $\wedge$ cleft(2,3)\end{tabular} &
\begin{tabular}[c]{@{}l@{}}Add a blue chair, a couch, and a red chair, \\ left to right, all under the TV.\end{tabular} \\ \hline
Fig. \ref{fig:pal1} & \begin{tabular}[c]{@{}l@{}}\# 0: a sink, 1: an oven.\\ type(2,"microwave") $\wedge$ type(3,"toaster") $\wedge$\\ property(2,"a blue microwave") $\wedge$\\ property(3,"a green toaster") $\wedge$\\ default(2) $\wedge$ default(3) $\wedge$\\ cright(2,1) $\wedge$ cleft(3,1) $\wedge$ cbelow(3,0)\end{tabular} &
\begin{tabular}[c]{@{}l@{}}Add a blue microwave right of the oven, \\ and a green toaster left of the oven and below the sink.\end{tabular} \\ \hline
\end{tabular}}
\caption{The logic formulations used for figures \ref{fig:pal1} with accompanying descriptive text. `\#' comments represent objects that were detected by the perception module and incorporated. Note the relation \texttt{default}: this sets width and height between 256 and 512 pixels to accommodate Stable Diffusion.}
\label{tab:nl_to_log_1}
\end{table}

\begin{table}[H]
\resizebox{\textwidth}{!}{%
\begin{tabular}{l|l|l}
\textbf{Figure} & \textbf{Logic Language} & \textbf{Natural Language} \\ \hline
Fig. \ref{fig:pal2} & 
\begin{tabular}[c]{@{}l@{}}type(0, "chair") $\wedge$ \\type(1, "couch") $\wedge$ \\type(2, "coffee table") $\wedge$\\ property(0, "a cozy brown leather chair") $\wedge$ \\property(1, "a black and white stripe couch") $\wedge$ \\property(2, "a stout coffee table") $\wedge$\\ default(0) $\wedge$ default(1) $\wedge$ default(2) $\wedge$\\ cleft(0, 1) $\wedge$\\ xeq(0, 1) $\wedge$\\ below(2, 0) $\wedge$\\ right(2, 0) $\wedge$\\ wider(1, 0) $\wedge$\\ wider(2, 0) $\wedge$\\ taller(1, 2)\end{tabular} & 
\begin{tabular}[c]{@{}l@{}}\textbf{Objects:}\\ A cozy brown leather chair,\\ a black and white stripe couch,\\ a stout coffee table in front of the couch.\\ \textbf{Constraints:}\\ The chair is completely left of the couch \\and horizontally aligned with it.\\ The coffee table is below the chair and to its right. \\The couch is wider than the chair, as is the coffee table. \\The couch is taller than the coffee table.\end{tabular} \\ \hline

Fig. \ref{fig:pal2} & 
\begin{tabular}[c]{@{}l@{}}type(0, "table") $\wedge$ \\type(1, "television") $\wedge$\\ property(0, \\\:"a big wooden kitchen table surrounded by chairs") $\wedge$ \\property(1, \\\:"a big flatscreen television mounted on a wall") $\wedge$\\ default(0) $\wedge$ default(1) $\wedge$\\ cbelow(0, 1) $\wedge$\\ cleft(1, 0) $\wedge$\\ wider\_value(0, 400) $\wedge$\\ taller\_value(0, 300) $\wedge$\\ wider\_value(1, 300)\end{tabular} & 
\begin{tabular}[c]{@{}l@{}}\textbf{Objects:} \\A big kitchen table surrounded by chairs, made of wood, \\a big flatscreen television mounted on a wall.\\ \textbf{Constraints:} \\The table is completely below the TV\\ and the TV is completely to the left of the table. \\The table is wider than 40\% of the image \\and taller than 30\% of the image. \\The TV is wider than 30\% of the image.\end{tabular} \\ \hline

Fig. \ref{fig:pal2} & 
\begin{tabular}[c]{@{}l@{}}type(0, "microwave") $\wedge$ \\type(1, "toaster") $\wedge$\\ property(0, "a brown microwave") $\wedge$ \\property(1, "a wood-panel pop-up toaster") $\wedge$\\ default(0) $\wedge$ default(1) $\wedge$\\ cleft(0, 1) $\wedge$\\ below(1, 0)\end{tabular} & 
\begin{tabular}[c]{@{}l@{}}\textbf{Objects:}\\ A brown microwave, \\a wood-panel pop-up toaster.\\ \textbf{Constraints:} \\The microwave is completely to the left of the toaster\\ and the toaster is below the microwave.\end{tabular} \\ \hline

Fig. \ref{fig:pal2} & 
\begin{tabular}[c]{@{}l@{}}type(0, "microwave") $\wedge$ \\type(1, "oven") $\wedge$\\ property(0, "a blue microwave") $\wedge$ \\property(1, "a white oven") $\wedge$\\ default(0) $\wedge$ default(1) $\wedge$\\ cleft(1, 0) $\wedge$\\ cbelow(1, 0) $\wedge$\\ right\_value(0, 500)\end{tabular} & 
\begin{tabular}[c]{@{}l@{}}\textbf{Objects:}\\ A blue microwave, a white oven.\\ \textbf{Constraints:}\\ The microwave is in the right 50\% of the image.\\ The oven is completely left and below the microwave.\end{tabular} \\ \hline
\end{tabular}}
\caption{The logic formulations used for figures \ref{fig:pal2} with accompanying descriptive text. Note the relation \texttt{default}: this sets width and height between 256 and 512 pixels to accommodate Stable Diffusion.}
\label{tab:nl_to_log_2}
\end{table}

\begin{table}[H]
\resizebox{\textwidth}{!}{%
\begin{tabular}{l|l|l}
\textbf{Figure} & \multicolumn{1}{c|}{\textbf{Logic Language}} &
  \multicolumn{1}{c}{\textbf{Natural Language}} \\ \hline
Fig. \ref{fig:comppal} & \begin{tabular}[c]{@{}l@{}}type(0,"microwave") $\wedge$ \\type(1,"oven") $\wedge$\\ property(0,"a blue microwave") $\wedge$\\ property(1,"a black oven") $\wedge$\\ default(0) $\wedge$ default(1) $\wedge$\\ cabove(0, 1)\end{tabular} & A blue microwave above a black oven. \\ \hline
Fig. \ref{fig:comppal} & \begin{tabular}[c]{@{}l@{}}type(0,"microwave") $\wedge$ \\type(1,"oven") $\wedge$ \\type(2,"refrigerator") $\wedge$\\ default(0) $\wedge$ default(1) $\wedge$ default(2) $\wedge$\\ cleft(2,1) $\wedge$ cright(0,1) $\wedge$ cabove(0,1)\end{tabular} &
\begin{tabular}[c]{@{}l@{}}A refrigerator left of an oven \\ and a microwave right and above the same oven.\end{tabular} \\ \hline
Fig. \ref{fig:comppal} & \begin{tabular}[c]{@{}l@{}}type(0, "microwave") $\wedge$ \\
type(1, "oven") $\wedge$ \\
type(2, "toaster") $\wedge$ \\
type(3, "sink") $\wedge$ \\
default(0) $\wedge$ default(1) $\wedge$ \\
default(2) $\wedge$ default(3) $\wedge$ \\
right(0, 1) $\wedge$ \\
above(0, 1) $\wedge$ \\
left(3, 1) $\wedge$ \\
above(3, 1) $\wedge$ \\
below(2, 0)\end{tabular} & 
\begin{tabular}[c]{@{}l@{}}A microwave, an oven, a toaster, and a sink. \\ The sink is left of and at least partly above the oven, \\ the microwave is right of and above the oven, \\ and the toaster is below the microwave.\end{tabular} \\ \hline
\end{tabular}}
\caption{The logic formulations used for figures \ref{fig:comppal} with accompanying descriptive text. Note the relation \texttt{default}: this sets width and height between 256 and 512 pixels to accommodate Stable Diffusion.}
\label{tab:nl_to_log_3}
\end{table}

\begin{table}[H]
\centering
{\fontsize{11pt}{12pt}\selectfont
\begin{tabular}{>{\raggedright\arraybackslash}p{0.7\linewidth}}
\hline
microwave, potted plant \\
microwave, plant \\
microwave, oven \\
toaster, plant \\
toaster, plant \\
couch, table \\
chair, table \\
plant, chair \\
couch, table \\
couch, tv \\
chair, plant, table \\
chair, table \\
couch, chair, plant \\
couch, chair, table, \\
chair, table \\
microwave, refrigerator \\
microwave, refrigerator, oven \\
refrigerator, microwave \\
oven, toaster, toaster \\
microwave, refrigerator, toaster \\
toaster, oven, refrigerator \\
plant, tv, table \\
plant, couch \\
chair, table \\
couch, tv, plant \\
refrigerator, refrigerator, oven \\
sink, toaster, microwave \\
sink, toaster, oven \\
microwave, microwave, oven \\
microwave, toaster \\
microwave, refrigerator, refrigerator \\
plant, couch, table \\
couch, chair, table \\
chair, table \\
chair, chair, table \\
oven, refrigerator, microwave \\
microwave, oven, refrigerator \\
oven, plant, \\
couch, couch, table \\
couch, chair \\
\hline
\end{tabular}
}
\caption{Objects generated for Figure \ref{fig:extrapal}, listed from top left to bottom right. These images do not contain positional constraints, and show that the SPRING's spatial reasoning module can provide good locations for objects even without human planning and closely tailored specifications.}
\label{tab:extrapal_objs}
\end{table}

\begin{figure}[H]
    \centering
    \includegraphics[width=\linewidth]{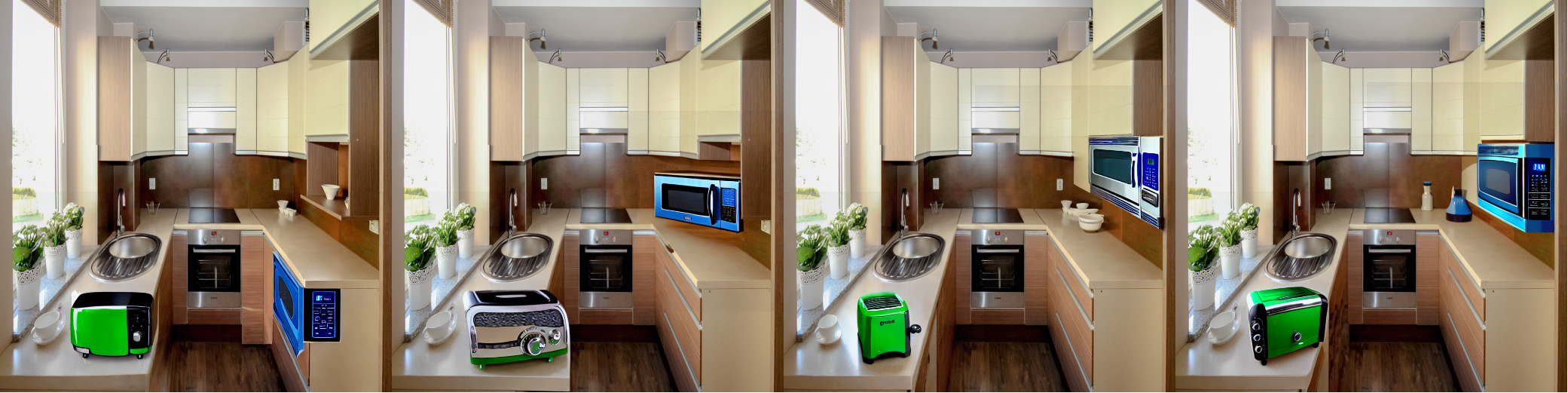}
    \caption{A closer look at some of the images generated for Figures \ref{fig:functionality} and \ref{fig:rollout}. Note that the background is minimally disturbed. Also note that, while the microwave can be anywhere right of the oven, it is placed in a reasonable place -- on the counter or built into the cabinets.}
    \label{fig:mainpic_extra}
\end{figure}

\section{Evaluating Spatial Reasoning Synthetic Scenarios}
\label{sec:syn_ds}

In this test, we compare the performance of the spatial reasoning module alone to a  generative adversarial network equipped with convex optimization as an embedded layer. We will evaluate each model's ability to meet both explicit constraints from the user and implicit preferences derived from the training data. The test scenarios will involve generating synthetic data for each object type, with the data distribution designed to express a strong preference for specific values. A range of common values for each preference will be established as acceptable, with values outside this range considered as not meeting the preference. The use of synthetic data is necessary as it allows us to know the acceptable range for each preference, which is not possible with real data. Both the spatial reasoning module and the GAN + CVX baseline will be trained on this preference-laden data, and then evaluated on their ability to satisfy constraints and preferences. The preference accuracies will be reported as the percentage of clauses that are satisfied.

The test includes three scenarios, each with specific data generation functions and acceptable preference ranges. The first scenario, ``basic'', involves objects with a higher likelihood of being generated on opposite halves of the image. The second scenario, ``tight'', has objects with specific width and height ratios with stricter acceptable ranges. The third scenario, ``complex'', has multiple preferences, such as object 1 being at least 1.5 times taller than its width, and object 2 having a specific y-coordinate value dependent on that of object 1. These scenarios demonstrate that the SRM's preference learning is effective and outperforms other methods such as GAN + CVX.

Two random functions were used to compose the synthetic datasets. \texttt{rnd(j, k)} randomly generates an integer between \textit{j} and \textit{k} from a normal distribution with mean $j + \frac{k-j}{2}$ and standard deviation $\frac{k-j}{12}$. Selected values are rounded, and constrained between j and k. For example, if object 1's \textit{x} value is drawn from \texttt{rnd(1, 500)}, then the mean of \textit{x} will be 250.5 and the standard deviation will be 41.583. \texttt{uni(j, k)} does the same with a uniform distribution between \textit{j} and \textit{k}.

The following sections include preference ranges (under ``checked preferences'') for each scenario. The preference accuracy is measured as a percentage of correctly satisfied clauses across 256 generated examples. The object associated with each variable is indicated in the subscript (e.g. $1 \leq x_{o1} \leq 500$ asserts that the x value of object 1 must be between 0 and 500 to satisfy the preference). Constraints are also given (under ``constraints'') in our constraint language. The constraint accuracy is measured as a percentage of correctly satisfied clauses across the same 256 generated examples.

\FloatBarrier
\subsection{Basic Scenario}
\FloatBarrier

\begin{table}[H]
\centering
{\fontsize{8pt}{12pt}\selectfont
\begin{tabular}{l|l|l|l|l}
\hline
\textbf{Object} & \textbf{x} & \textbf{y} & \textbf{width (w)} & \textbf{height (h)} \\\hline
1 & \texttt{rnd(1,500)} & \texttt{uni(400, 550)} & \texttt{rnd(192,256)} & \texttt{rnd(128,256)} \\
2 & \texttt{rnd(500,1000)} & \texttt{uni(400, 550)} & \texttt{rnd(192,256)} & \texttt{rnd(128,256)}
\end{tabular}
}
\end{table}

\textbf{Constraints:}
\begin{itemize}
\scriptsize
\item above(o1, o2, 300) -- Object 1 is above object 2 by at least 300 per mille.
\end{itemize}

\textbf{Checked preferences:}
\begin{itemize}
\scriptsize
\item $1 \leq x_{o1} \leq 500$
\item $500 \leq x_{o2} \leq 1000$
\end{itemize}

\FloatBarrier
\subsection{Tight Scenario}
\FloatBarrier

\begin{table}[H]
\centering
{\fontsize{8pt}{12pt}\selectfont
\begin{tabular}{l|l|l|l|l}
\hline
\textbf{Object} & \textbf{x} & \textbf{y} & \textbf{width (w)} & \textbf{height (h)} \\\hline
1 & \texttt{rnd(1,1000)} & \texttt{rnd(300,700)} & \texttt{rnd(220,256)} & \texttt{rnd(120,150)} \\
2 & \texttt{rnd(1,1000)} & \texttt{rnd(300,700)} & \texttt{rnd(120,150)} & \texttt{rnd(120,150)} \\ 
3 & \texttt{rnd(1,1000)} & \texttt{rnd(300,700)} & \texttt{rnd(120,150)} & \texttt{rnd(220,256)} \\ 
\end{tabular}
}
\end{table}

\textbf{Constraints:}
\begin{itemize}
\scriptsize
\item left(o1, o2, 180) -- Object 1 is left of object 2 by at least 180 per mille.
\item left(o2, o3, 180) -- Object 2 is left of object 3 by at least 180 per mille.
\end{itemize}

\textbf{Checked preferences:}
\begin{multicols}{3}
\scriptsize
\begin{itemize}
\item $220 \leq w_{o1} \leq 256$
\item $120 \leq w_{o2} \leq 150$
\item $120 \leq w_{o3} \leq 150$
\item $120 \leq h_{o1} \leq 150$
\item $120 \leq h_{o2} \leq 150$
\item $220 \leq h_{o3} \leq 256$
\end{itemize}
\end{multicols}

\FloatBarrier
\subsection{Complex Scenario}
\FloatBarrier

\begin{table}[H]
\centering
{\fontsize{8pt}{12pt}\selectfont
\begin{tabular}{l|l|l|l|l}
\hline
\textbf{Object} & \textbf{x} & \textbf{y} & \textbf{width (w)} & \textbf{height (h)}  \\\hline
1 & \texttt{rnd(1,1050)} & \texttt{rnd(375,565)} & \texttt{rnd(64, 128)} & \texttt{rnd($w_{o1}\times1.5$,200)}  \\
2 & \texttt{rnd(1,1050)} & \texttt{rnd($y_{o1}-10$,$y_{o1}+10$)} & \texttt{rnd(64, 128)} & \texttt{rnd($w_{o2}\times1.5$,200)}  \\ 
3 & \texttt{rnd(1,900)} & \texttt{rnd(1,144)} & \texttt{rnd(64, 128)} & \texttt{rnd($w_{o3}\times2$,$w_{o3}\times2$)} \\ 
4 & \texttt{rnd(1,1050)} & \texttt{rnd(400,665)} & \texttt{rnd(64, 128)} & \texttt{rnd($w_{o4}-10$,$w_{o4}+10$)} \\ 
\end{tabular}
}
\end{table}

\textbf{Constraints:}
\begin{itemize}
\scriptsize
\item left(o1, o2, 400) -- Object 1 is left of object 2 by at least 400 per mille.
\item above(o2, o4, 200) -- Object 2 is above object 4 by at least 200 per mille.
\item right(o4, o1, 250) -- Object 4 is right of object 1 by at least 250 per mille.
\item right\_value(o1, 500) -- Object 1 is in the right 50\% of the image.
\item wider\_value(o4, 250) -- Object 4 is at least 25\% the width of the image.
\item above\_value(o3, 250) -- Object 2 is in the top 25\% of the image.

\end{itemize}

\textbf{Checked preferences:}
\begin{multicols}{3}
\scriptsize
\begin{itemize}
\item $1 \leq x_{o1} \leq 1050$
\item $1 \leq x_{o2} \leq 1050$
\item $1 \leq x_{o3} \leq 900$
\item $1 \leq x_{o4} \leq 1050$
\item $375 \leq y_{o1} \leq 565$
\item $y_{o1} - 10 \leq y_{o2} \leq y_{o1} + 10$
\item $1 \leq y_{o3} \leq 144$
\item $400 \leq y_{o4} \leq 665$
\item $64 \leq w_{o1} \leq 128$
\item $64 \leq w_{o2} \leq 128$
\item $64 \leq w_{o3} \leq 128$
\item $64 \leq w_{o4} \leq 128$
\item $w_{o1} \times 1.5 \leq h_{o1} \leq 200$
\item $w_{o2} \times 1.5 \leq h_{o2} \leq 200$
\item $w_{o3} \times 2   \leq h_{o3} \leq w_{o3} \times 2$
\item $w_{o4} - 10 \leq h_{o4} \leq w_{o4} + 10$
\end{itemize}
\end{multicols}




\end{document}